\documentclass[letterpaper]{article}
\usepackage{iMacHSR}
\usepackage{times}
\usepackage{helvet}
\usepackage{courier}
\usepackage[hyphens]{url}
\usepackage{graphicx}
\usepackage{natbib}
\usepackage{caption}
\usepackage{algorithm}
\usepackage{algorithmic}
\usepackage{amsmath}
\usepackage{cleveref}
\usepackage{tabularx}
\usepackage[caption=true,font=normalsize,labelfont=sf,textfont=sf]{subfig}
\usepackage{multirow}
\usepackage{amsfonts}
\usepackage{float}
\usepackage{afterpage}
\usepackage{pifont}

\usepackage{newfloat}
\usepackage{listings}
\DeclareCaptionStyle{ruled}{labelfont=normalfont,labelsep=colon,strut=off}
\floatstyle{ruled}
\newfloat{listing}{tb}{lst}{}
\floatname{listing}{Listing}

\Crefname{figure}{Fig.}{Figs.}
\Crefname{equation}{Eq.}{Eqs.}
\Crefname{section}{Sec.}{Secs.}
\Crefname{subsection}{Sec.}{Secs.}
\crefname{subsection}{Sec.}{Secs.}
\crefname{section}{Sec.}{Secs.}

\def\eg{\emph{e.g.}}

\def\ie{\emph{i.e.}}

\newcommand{\verticaltext}[2][0pt]{%
  \raisebox{#1}{\parbox[t]{1em}{\rotatebox[origin=c]{90}{#2}}}%
}

\newcommand{\xmark}{\ding{55}}

\newtheorem{theorem}{Theorem}

\title{iMacHSR: Intermediate Multi-Access Heterogeneous Supervision and Regularization Scheme Toward Architecture-Agnostic Training}
\author{
    Wei-Bin Kou\textsuperscript{\rm 1, \rm 2},
    Guangxu Zhu\textsuperscript{\rm 3},
    Yichen Jin\textsuperscript{\rm 1},
    Bingyang Cheng\textsuperscript{\rm 1}, \\
    Shuai Wang\textsuperscript{\rm 4},
    Ming Tang\textsuperscript{\rm 2},
    Yik-Chung Wu\textsuperscript{\rm 1}
}
\affiliations{
    \textsuperscript{\rm 1}Department of Electrical and Electronic Engineering, The University of Hong Kong, Hong Kong, China.\\
    \textsuperscript{\rm 2}Department of Computer Science and Engineering, Southern University of Science and Technology, Shenzhen, China.\\
    \textsuperscript{\rm 3}Shenzhen International Center For Industrial And Applied Mathematics, Shenzhen Research Institute of Big Data, \\Shenzhen, China.\\
    \textsuperscript{\rm 4}Shenzhen Institutes of Advanced Technology, Chinese Academy of Sciences, Shenzhen, China.\\
}

\begin{document}

\maketitle

\begin{abstract}
While deep supervision is a powerful training strategy by supervising intermediate layers with auxiliary losses, it faces three underexplored problems: (I) Existing deep supervision techniques are generally bond with specific model architectures strictly, lacking generality. (II) The identical loss function for intermediate and output layers causes intermediate layers to prioritize output-specific features prematurely, limiting generalizable representations. (III) Lacking regularization on hidden activations risks overconfident predictions, reducing generalization to unseen scenarios. To tackle these challenges, we propose an architecture-agnostic, \underline{i}ntermediate \underline{M}ulti-\underline{ac}cess \underline{H}eterogeneous \underline{S}upervision and \underline{R}egularization (iMacHSR) scheme. Specifically, the proposed iMacHSR introduces below integral strategies: (I) we select multiple intermediate layers based on predefined architecture-agnostic standards; (II) loss functions (different from output-layer loss) are applied to those selected intermediate layers, which can guide intermediate layers to learn diverse and hierarchical representations; and (III) negative entropy regularization on selected layers' hidden features discourages overconfident predictions and mitigates overfitting. These intermediate terms are combined into the output-layer training loss to form a unified optimization objective, enabling comprehensive optimization across the network hierarchy. We then take the semantic understanding task as an example to assess iMacHSR and apply iMacHSR to several model architectures. Extensive experiments on multiple datasets demonstrate that iMacHSR outperforms conventional output-layer single-point supervision method up to 9.19\% in mIoU. 
\end{abstract}

\section{INTRODUCTION}
Recent advances in deep learning (DL) have propelled prediction accuracy to unprecedented levels \cite{yang2025mtl,xu2024sctnet,esmaeilpour2022zero,chen2017deeplabsemanticimagesegmentation,wang2020deephighresolutionrepresentationlearning,yu2018bisenet}. However, as DL model depth increases, traditional output-layer single-point supervision training method faces inherent limitations, such as gradient vanishing \cite{9526915,hanin2018neural,guo2024take,kera2020gradient}, under-optimized intermediate layers \cite{hao2020labelenc,liu2024detection}, etc. These limitations often degrade the potential of deep architectures. To mitigate such limitations, current explorations primarily focus on architectural innovations, such as residual connections \cite{he2016deep,tang2024boosting,li2023residual,kong2022residual}, attention mechanisms \cite{liu2021adaattn,islam2021hybrid,han2024agent,yu2024nonlocal}, etc. Yet, they overlook the critical role of supervision on intermediate layers. Without explicit supervision at intermediate stages, these layers may fail to learn task-relevant features, leading to performance plateaus. 

Deep supervision \cite{lee2015deeply,ren2025deepmim,zhang2022contrastive,8099532,pang2021toward} supplements this absence of supervision on hidden layers. Specifically, deep supervision applies auxiliary losses to intermediate layers of a network, in addition to the output-layer loss. This provides explicit supervision to learn the intermediate feature representations. However, three under-investigated issues exist in deep supervision: (I) existing deep supervision techniques are typically tightly coupled with specific model architectures, limiting their generality. For example, ICNet \cite{Zhao_2018_ECCV} applies auxiliary losses to low-resolution intermediate predictions in a cascaded framework. (II) the loss functions applied to the intermediate layers are identical to that of output layer. This causes the intermediate layers to prioritize output-specific features prematurely, limiting to learn generalizable representations. (III) the absence of regularization on hidden activations risks overconfident predictions, reducing model generalization to unseen scenarios.

\begin{figure*}[!t]
\includegraphics[width=\linewidth]{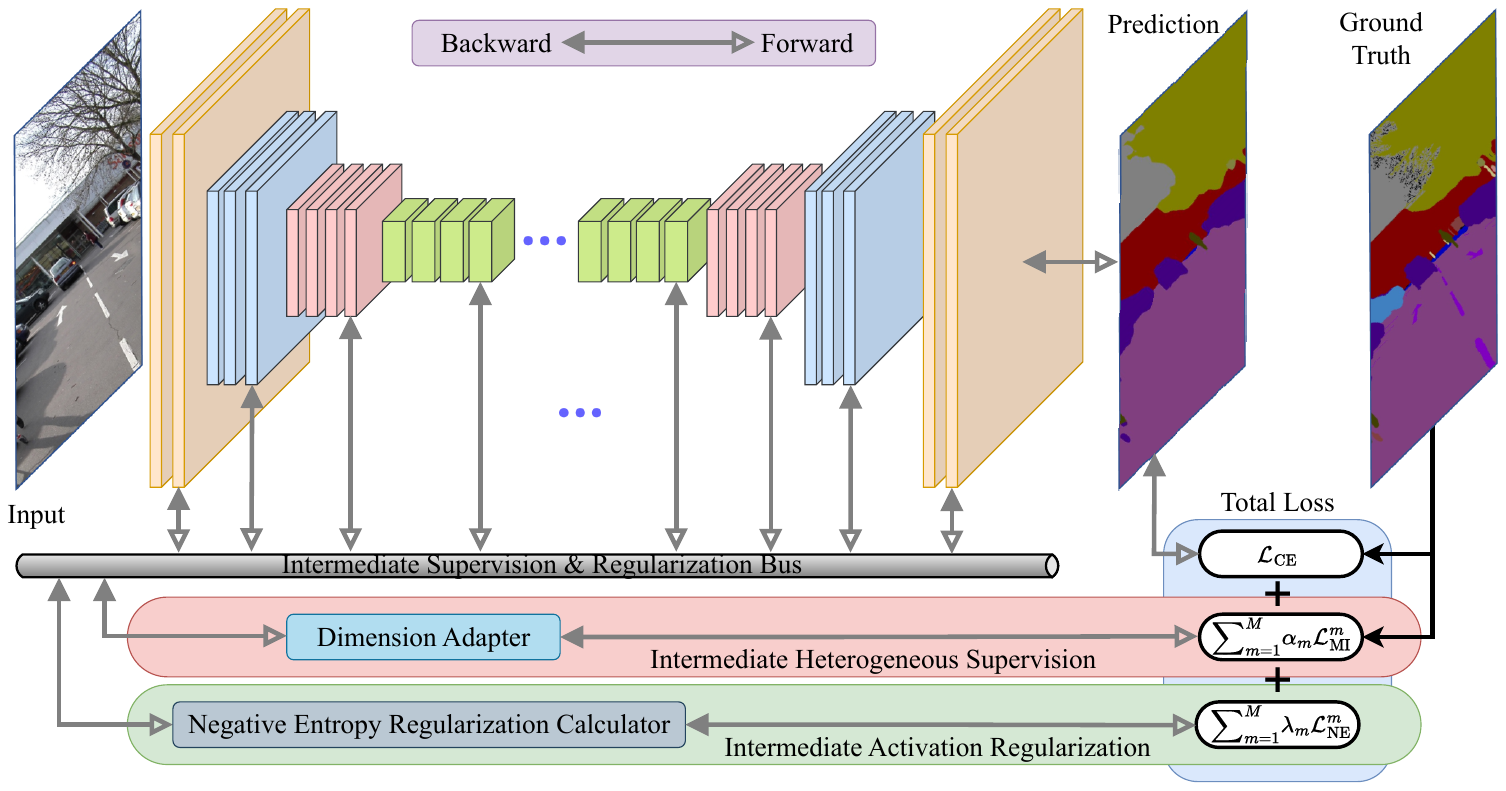}
\caption{Overview of the proposed iMacHSR training scheme, taking semantic segmentation task as an example.}
\label{fig:iMacRS_framework}
\end{figure*}

To bridge these gaps, we propose \underline{i}ntermediate \underline{M}ulti-\underline{ac}cess \underline{H}eterogeneous \underline{S}upervision and \underline{R}egularization (iMacHSR), a model architecture-agnostic training scheme that integrates different losses (from output-layer loss) and regularization on multiple selected intermediate layers, in addition to the output-layer supervision loss. Specifically, iMacSR adopts following three integral policies: (I) Intermediate Point Selection: we select multiple intermediate layers based on predefined architecture-agnostic criteria to ensure flexibility across various model designs. For example, we can choose layers at key transition points in the network, such as layers between major blocks (e.g., ResNet stages or transformer layers). (II) Heterogeneous Losses: different losses are applied to intermediate and output layers, which guides intermediate layers to focus on learning diverse and hierarchical representations. For instance, for segmentation task, different from output layer's cross entropy loss, mutual information between latent features and ground truth could be used as intermediate loss. (III) Negative Entropy Regularization: we propose to calculate negative entropy on each intermediate point's latent features as a regularizer. This helps to improve generalization by penalizing overconfident hidden feature predictions. In addition, we carry out a theoretical convergence analysis for iMacHSR, providing insights into how the proposed iMacHSR impacts the convergence of DL model. We also analyze the time and space complexity of iMacHSR, revealing that iMacHSR incurs linear overheads to the number of intermediate points, in both time and space complexity. The proposed iMacHSR is illustrated in \Cref{fig:iMacRS_framework}. We evaluate iMacHSR based on semantic segmentation task and apply iMacHSR to multiple model architectures. Extensive Experiments on Cityscapes \cite{Cordts2016Cityscapes}, CamVid \cite{brostow2008segmentation}, and SynthiaSF \cite{ros2016synthia} datasets demonstrate that iMacHSR-trained model achieves 9.19\% higher mIoU than that of conventional output-layer supervision training approach. 

The main contributions are highlighted as follows:
\begin{enumerate}
    \item Deep supervision faces challenges such as the dependence on specific model architecture, the learned premature intermediate features, and the absence of regularization for intermediate activations. To mitigate these issues, we propose iMacHSR, a model architecture-independent training scheme that integrates both heterogeneous losses and regularization on multiple intermediate layers. 
    \item We additionally conduct theoretical convergence analysis for iMacHSR, which suggests that iMacHSR holds $\mathcal{O}(1/\sqrt{T})$ convergence rate that matches standard SGD optimization, proving iMacHSR does not harm asymptotic convergence. 
    \item We also analyze time and space complexity of iMacHSR, indicating that iMacHSR introduces linear overheads with respect to the number of the intermediate points, for both time and space complexity.
    \item We use the semantic segmentation task as an example to assess iMacHSR and apply it to multiple model architectures. Extensive experiments on multiple datasets demonstrate that iMacHSR outperforms conventional output-layer supervision training method up to 9.19\% in mIoU. In addition, we conduct ablation studies to explore how the number of intermediate points, the distance between adjacent intermediate points, and the positions of the intermediate points affect the model performance.
\end{enumerate}

\section{Related Work}
\label{related_work}
\subsection{DL Model Optimization}
DL model optimization \cite{yao2021adahessian,sankaran2021deeplite,park2023self,mallik2023priorband} includes a suite of algorithms and techniques to optimize a loss function across DL model parameters. Stochastic gradient descent (SGD) lays the groundwork and back-propagation provides a computationally feasible way for training DL models \cite{amari1993backpropagation,refinetti2023neural,shumailov2021manipulating,tang2024dp,li2024backpropagation,kirsch2021meta}. Based on these two elements, innovations such as Adam \cite{kingma2014adam}, RMSprop \cite{hinton2012neural}, and AdaGrad \cite{duchi2011adaptive} later emerged. They offer adaptive learning rates that can resolve some limitations found in SGD, particularly in terms of convergence and stability across various DL model architectures. DL model optimization is also closely linked with techniques intended to improve the generalization and stability of DL models. Regularization strategies such as dropout \cite{hinton2012improving}, L1/L2 regularization \cite{tibshirani1996regression,hoerl1970ridge}, etc., are critical in preventing overfitting and ensuring robust model performance. Similarly, normalization techniques like batch normalization \cite{ioffe2015batch} and layer normalization \cite{ba2016layer} have been pivotal in stabilizing training. Despite significant advancements, DL model optimization continues to face challenges, such as gradient vanishing \cite{hanin2018neural,guo2024take}, under-optimized hidden features \cite{hao2020labelenc}, particularly in training extremely deep networks. In order to mitigate these problems, this work supplements deep supervision's weaknesses to present iMacHSR that introduces intermediate multi-point heterogeneous supervision and regularization.

\subsection{Deep Supervision}
Deep supervision~\cite{lee2015deeply,zhang2022contrastive,li2022comprehensive} has been previously explored as a method to aid the training of deep networks, potentially addressing gradient vanishing issues~\cite{hochreiter2001gradient}. For example, GoogleNet~\cite{szegedy2015going} incorporates two additional supervision layers at intermediate stages. DSN~\cite{wang2015training} introduces auxiliary supervision branches at specific intermediate layers. PSPNet \cite{Zhao_2017_CVPR} incorporates an auxiliary classifier to calculate the pixel-wise cross-entropy between the auxiliary predictions and the ground truth. BiSeNet \cite{yu2018bisenet} applies deep supervision to a spatial path and a context path to ensure balance between spatial detail and global context. Gated-SCNN \cite{9009833} introduces shape-based intermediate losses to enhance the learning of shape-aware features. ICNet \cite{Zhao_2018_ECCV} uses deep supervision by adding auxiliary loss branches to low-resolution intermediate predictions in a cascaded framework. With the advent of techniques like batch normalization~\cite{ioffe2015batch} and residual learning~\cite{he2016deep}, gradient vanishing problem has become less common, which may explain the reduced focus on deep supervision in recent years. While deep supervision has broad applications, it encounters challenges such as the dependence on specific model architecture, prioritizing output-specific intermediate features prematurely, and the lack of regularization for intermediate activations, etc. This paper presents iMacHSR to address these issues.

\section{Methodology}
\label{methodology}
We firstly elaborate the proposed iMacHSR. We then conduct convergence analysis for iMacHSR. Finally, we discuss the time and space complexity of iMacHSR. 

\subsection{The Proposed iMacHSR}
\label{iMacSR_formulation}
The key notations in iMacHSR formulation are summarized in \Cref{tab:iMacSR_notations}.
Let $\mathcal{D} = \{(x_i, y_i)\}_{i=1}^{|\mathcal{D}|}$ denote the training dataset, where $x_i$ is an input image and $y_i$ is its ground truth. In addition, we set $M$ intermediate supervision and regularization points for the network $\theta$, and such points are denoted as $\{G_1, \dots, G_M\}$. There exist consecutive layers between two adjacent intermediate points. 

\begin{table}[t]
    \centering
    \renewcommand{\arraystretch}{1.0}
    \setlength{\tabcolsep}{4.0pt}
    \begin{tabularx}{\linewidth}{ll}
    \hline
        \textbf{Symbols} & \textbf{Definitions} \\ \hline
        $\mathcal{D}$ & Training dataset \\ 
        $(x_i, y_i)$ & Input image and the corresponding ground truth \\ 
        $\theta$ & DL model parameters \\ 
        $M$ & Total number of intermediate points \\ 
        $G_m$ & Intermediate point $m$ \\ 
        $z^m$ & Latent feature maps at point $G_m$ \\ 
        $\mathcal{L}_{\text{CE}}$ & Cross-entropy loss \\ 
        $\mathcal{L}_{\text{MI}}^m$ & Mutual information loss for point $G_m$ \\ 
        $\mathcal{L}_{\text{NE}}^m$ & Negative entropy regularization for point $G_m$ \\ 
        $\alpha_m, \lambda_m$ & Loss weights for point $G_m$ \\ 
        \hline
    \end{tabularx}
\caption{Key Notations of iMacHSR Formulation}
\label{tab:iMacSR_notations}
\end{table}

For the proposed iMacHSR, we firstly select some intermediate points based on some predefined model architecture-agnostic rules. Specifically, we propose to choose layers at key transition points in the network, as these points often represent significant changes in feature representation or abstraction. Examples include but not limit to:  
\begin{itemize}
    \item \textbf{Before or after downsampling} (e.g., pooling layers) to capture changes in spatial size and feature granularity.  
    \item \textbf{Between major blocks} (e.g., ResNet stages or transformer layers) to leverage the differences in feature abstraction between hierarchical stages.  
    \item \textbf{At bottleneck layers}, where the feature dimensions are compressed, highlighting critical information.  
    \item \textbf{Before or after attention mechanisms} to capture how signal is distributed or aggregated across feature maps.
    \item \textbf{At skip connections in encoder-decoder architectures} (e.g., UNet) to include both high-resolution and low-resolution contextual information.  
    \item \textbf{Near activation function changes or normalization layers}, where feature transformations can significantly influence downstream learning.  
    \item \textbf{At feature fusion points} in multi-branch architectures, to capture the integration of diverse feature streams.  
\end{itemize}  

These transition points provide a comprehensive view of how features evolve throughout the network, enabling more effective supervision and training.

We then propose to impose supervision and regularization on those selected intermediate layers. Therefore, the proposed iMacHSR's optimization objective is three-fold:
\begin{itemize}
    \item Conventional output-layer loss: In most classification and segmentation task, cross entropy (CE) loss (denoted as $\mathcal{L}_{\text{CE}}$) is used as optimization objective. 

\item Intermediate heterogeneous loss: 
For point $G_m$, the latent feature for the $x_i$ is $z_i^m = \theta^{G_m}(x_i)$. We maximize mutual information between $z^m$ and labels $y$ via
\begin{equation}
\mathcal{L}_{\text{MI}}^m \!=\! 1/|\mathcal{D}| \sum\nolimits_{(x_i, y_i) \in \mathcal{D}} \!L_{MI}(q_m(z_i^m; \phi_m), y_i; \theta),
\label{eq:mutual_info}
\end{equation}
where $L_{MI}(\cdot)$ is the image-wise mutual information (MI) loss, $q_m(\cdot; \phi_m)$ is a dimension adapter with parameters $\phi_m$ for point $G_m$, aligning the latent features' dimension with ground truth's dimension for calculating MI. By focusing on the shared information between intermediate features and the ground truth, MI enables the model to learn representations that are both meaningful and discriminative, often leading to improved performance and generalization.

\item Intermediate negative entropy (NE) regularization:
For point $G_m$, to prevent overconfidence of feature representation, we minimize negative entropy of $z^m$ to encourage the model to be more uncertain about its predictions. NE regularizer is formulated as
\begin{equation}
\mathcal{L}_{\text{NE}}^m = 1/|\mathcal{D}| \sum\nolimits_{(x_i, y_i) \in \mathcal{D}} L_{NE}(z_i^m; \theta),
\label{eq:negative_entropy}
\end{equation}
where $L_{NE}(\cdot)$ means image-wise negative entropy regularization loss.
\end{itemize}

In summary, the total optimization objective is
\begin{equation}
\mathcal{L}_{\text{T}} = \mathcal{L}_{\text{CE}} + \sum\nolimits_{m=1}^M \left( \alpha_m \mathcal{L}_{\text{MI}}^m + \lambda_m \mathcal{L}_{\text{NE}}^m \right),
\label{eq:total_loss_flex}
\end{equation}
where $\alpha_m, \lambda_m$ are coefficients of supervision loss and regularization term, respectively, for intermediate point $G_m$.

As usual, the proposed iMacHSR optimizes the DL model via gradient descent for multiple rounds until convergence. For each round, it follows below steps: (I) Forward Pass: For an input image $x_i$, it computes features $\{z_i^1, \dots, z_i^M\}$ at each point $G_m$ and the final prediction $\hat{y}_i$. (II) Loss Computation: It calculates the total loss $\mathcal{L}_{\text{T}}$ using \Cref{eq:total_loss_flex}, which includes $\mathcal{L}_{\text{CE}}$, $\{\mathcal{L}_{\text{MI}}^m\}_{m=1}^{m=M}$, and $\{\mathcal{L}_{\text{NE}}^m\}_{m=1}^{m=M}$. (III) Back Propagation: It computes gradients of $\mathcal{L}_{\text{T}}$ with respect to DL model parameters $\theta$ and auxiliary dimension adapter $\{\phi_m\}$. (IV) Parameter Update: It updates $\theta$ and $\{\phi_m\}_{m=1}^{m=M}$ using the Adam optimizer \cite{kingma2014adam}, \ie,
\begin{equation}
    \theta \leftarrow \theta - \eta \nabla_\theta \mathcal{L}_{\text{T}}, \quad \phi_m \leftarrow \phi_m - \eta \nabla_{\phi_m} \mathcal{L}_{\text{T}},
\end{equation}
where $\eta$ is the learning rate.

\begin{algorithm}[tp]
\caption{iMacHSR}
\label{iMacRS_algo}
\begin{algorithmic}[1]
\REQUIRE Training dataset $\mathcal{D}$, model $\theta$ with intermediate points $\{G_1, \dots, G_M\}$, learning rate $\eta$, epochs $T$
\ENSURE Trained model ${\theta^*}$

\STATE Initialize $\theta$ with $\theta_0$, auxiliary dimension adapter $\{\phi_1, \dots, \phi_M\}$, weights $\{\alpha_1, \lambda_1, \dots, \alpha_M, \lambda_M\}$

\FOR{epoch $= 1$ to $T$}
    \FOR{each batch $(x_i, y_i) \in \mathcal{D}$}
        \STATE \textbf{Forward Pass:}
        \STATE $\hat{y}_i, \{z_i^1, \dots, z_i^M\} \gets \theta(x_i)$

        \STATE \textbf{Loss Computation:}
        \STATE $\mathcal{L}_{\text{CE}} \gets \{(y_i, \hat{y}_i)\}_{i=1}^{|\mathcal{D}|}$ 
        \FOR{$m = 1$ to $M$}
            \STATE $\mathcal{L}_{\text{MI}}^m \gets$ \Cref{eq:mutual_info},~~ $\mathcal{L}_{\text{NE}}^m \gets$ \Cref{eq:negative_entropy}
        \ENDFOR
        \STATE $\mathcal{L}_{\text{T}} \gets$ \Cref{eq:total_loss_flex}

        \STATE \textbf{Back Propagation \& Update:}
        \STATE Compute $\nabla_\theta \mathcal{L}_{\text{T}}$, $\nabla_{\phi_m} \mathcal{L}_{\text{T}}$ for all $m$
        \FOR{$m = 1$ to $M$}
            \STATE $\phi_m \gets \phi_m - \eta \nabla_{\phi_m} \mathcal{L}_{\text{T}}$
        \ENDFOR
        \STATE $\theta \gets \theta - \eta \nabla_\theta \mathcal{L}_{\text{T}}$
    \ENDFOR
\ENDFOR

\RETURN ${\theta^*}$
\end{algorithmic}
\end{algorithm}

In conclusion, iMacHSR is outlined in \Cref{iMacRS_algo}.

\subsection{Convergence Analysis of iMacHSR}
\label{iMacSR_convergence}

To clearly conduct convergence analysis, some assumptions are made. Specifically, for each component $s \in \{\text{CE}, \{_\text{MI}^m\}, \{_\text{NE}^m\}\}$, $\mathcal{L}_s$ satisfies: 
\begin{itemize}
    \item L-smoothness: There exists $L_s > 0$ such that $\forall \theta, \theta'$, $\|\nabla\mathcal{L}_s(\theta) - \nabla\mathcal{L}_s(\theta')\| \leq L_s\|\theta - \theta'\|;$
    
    \item Bounded Gradients: There exists $G_s > 0$ such that $\forall \theta$, $\mathbb{E}[\|\nabla\mathcal{L}_s(\theta)\|^2] \leq (G_s)^2;$
    
    \item Bounded Variance: There exists $(\sigma_s)^2 > 0$ such that $\forall \theta$, $\mathbb{E}[\|\nabla\mathcal{L}_s(\theta) - \nabla\mathcal{L}_s(\theta)\|^2] \leq (\sigma_s)^2.$
\end{itemize}
Based on these assumptions, we can conclude below Theorem 1 about the convergence rate of the proposed iMacHSR.

\begin{theorem}
Let $L_{\text{max}} = \max(L_{\text{CE}}, \alpha_mL_{\text{MI}}^m, \lambda_mL_{\text{NE}}^m)$, $G_{\text{T}}^2 = G_{\text{CE}}^2 + \sum_{m=1}^M (\alpha_m^2(G^{m}_{\text{MI}})^2 + \lambda_m^2(G^{m}_{\text{NE}})^2)$, and $\sigma_\text{T}^2= \sigma_{\text{CE}}^2 + \sum_{m=1}^M(\alpha_m^2(\sigma^m_{\text{MI}})^2 + \lambda_m^2(\sigma^m_{\text{NE}})^2)$. After $T$ iterations of training with $\eta_t = \frac{\eta}{\sqrt{T}}$, we have
\begin{align}
    \frac{1}{T}\sum_{t=1}^T \mathbb{E}\|\nabla\mathcal{L}_{\text{T}}(\theta_t)\|^2 \leq \underbrace{\frac{2\Delta}{{\eta}\sqrt{T}}}_{\text{Initial gap}} + \underbrace{\frac{L_{\text{max}}\eta}{\sqrt{T}}\left(G_{\text{T}}^2 + \sigma_\text{T}^2\right)}_{\text{Variance terms}},
    \label{theorem_1}
\end{align}
where $\Delta = \mathcal{L}_{\text{T}}(\theta_0) - \mathcal{L}_{\text{T}}^*$, $\theta_0$ is the initial model parameters, $\mathcal{L}_{\text{T}}^*$ is the theoretical optimal loss.
\end{theorem}

From Theorem 1, we can conclude following insights: (I) The $\mathcal{O}(1/\sqrt{T})$ rate matches standard non-convex SGD, proving iMacHSR does not harm asymptotic convergence. (II) The gradient bound is positively related to the number of  intermediate points (\ie, $M$), which is controllable via the number selection of intermediate point (e.g., $M = \mathcal{O}(\log D)$ for DL model depth $D$). 

This convergence theorem is proven in \textit{Appendix I of Supplementary Materials}.

\subsection{Complexity Analysis of iMacHSR}
\label{iMacSR_complexity}
To clearly conduct complexity analysis, we denote some notations as follows: $B$ is the batch size,$D$ is the depth of the DL model, $W$ and $H$ are the width and the height of the input image, $w_m\,\ h_m, \text{and}\ C_m$ are the width, the height, and the channel number of the latent feature maps at intermediate point $G_m$. Notably, we just offer the complexity results in following parts, and the detailed derivation process can be viewed in \textit{Appendix II of Supplementary Materials}.

\subsubsection{Time Complexity.}
For each batch, the time is composed of three parts: forward time, loss computation time, and backward time. Specifically, the forward time is $O(D+M)$; the loss computation time is $O(B((M+1)WHK+\sum_{m=1}^M Bw_mh_mC_m))$; the backward time is $O(D)$. In conclusion, the total time of each batch is $O(D+M) + O(B((M+1)WHK+\sum_{m=1}^M Bw_mh_mC_m)) + O(D)$.

\begin{table*}[tp]
\centering
\setlength{\tabcolsep}{3.3pt}
\begin{tabularx}{\linewidth}{c|c|c|cccc|cccc|cccc}
\hline
\multirow{2}{*}{Models} &\multirow{2}{*}{Backbone}    & \multirow{2}{*}{iMacRS?} & \multicolumn{4}{c|}{Cityscapes Dataset (\%)}                        & \multicolumn{4}{c|}{CamVid Dataset (\%)}                           & \multicolumn{4}{c}{SynthiaSF Dataset (\%)}                        \\ \cline{4-7} \cline{8-11} \cline{12-15} 
                            &            &                       & mIoU           & mF1            & mPre     & mRec        & mIoU           & mF1            & mPre     & mRec        & mIoU           & mF1            & mPre     & mRec        \\ \hline
\multirow{2}{*}{DeepLabv3+} &\multirow{2}{*}{ResNet18} & \xmark                      & 43.76          & 50.40          & 51.54          & 50.77          & 76.02          & 82.43          & 83.07          & 82.46          & 33.28          & 37.27          & 38.98          & 36.45          \\ 
                          &  & \checkmark                     & \textbf{47.78} & \textbf{56.28} & \textbf{59.64} & \textbf{55.64} & \textbf{76.13} & \textbf{82.52} & \textbf{83.09} & \textbf{82.57} & \textbf{34.28} & \textbf{39.13} & \textbf{42.74} & \textbf{37.60} \\ \hline
\multirow{2}{*}{SeaFormer} &\multirow{2}{*}{-}  & \xmark                      & 27.40          & 30.99          & 30.55          & 32.14          & 50.69          & 56.00          & 55.40          & 56.89          & \textbf{24.74} & \textbf{29.70} & \textbf{32.68} & \textbf{29.19} \\
                          &  & \checkmark                     & \textbf{29.82} & \textbf{34.19} & \textbf{33.80} & \textbf{35.45} & \textbf{55.83} & \textbf{62.39} & \textbf{64.19} & \textbf{62.54}  & 24.20          & 29.23          & 32.20          & 29.00          \\ \hline
\multirow{2}{*}{TopFormer} &\multirow{2}{*}{-}  & \xmark                      & 32.76          & 37.64          & 36.92          & 39.24          & 63.10          & 70.22          & 71.88          & 70.25          & 28.37          & 33.75          & 36.97          & 32.99          \\
                         &   & \checkmark                     & \textbf{34.28} & \textbf{39.96} & \textbf{40.41} & \textbf{40.60} & \textbf{66.38} & \textbf{74.50} & \textbf{77.47} & \textbf{73.60} & \textbf{28.70} & \textbf{34.04} & \textbf{37.22} & \textbf{33.20} \\ \hline
\end{tabularx}
\caption{The quantitative performance comparison of enabling iMacHSR against disabling iMacHSR for multiple models}
\label{tab:iMacRS_quantitative_comp}
\end{table*}

\begin{figure*}[tp]
\centering
\subfloat[\footnotesize mIoU]{\includegraphics[width=0.24\linewidth]{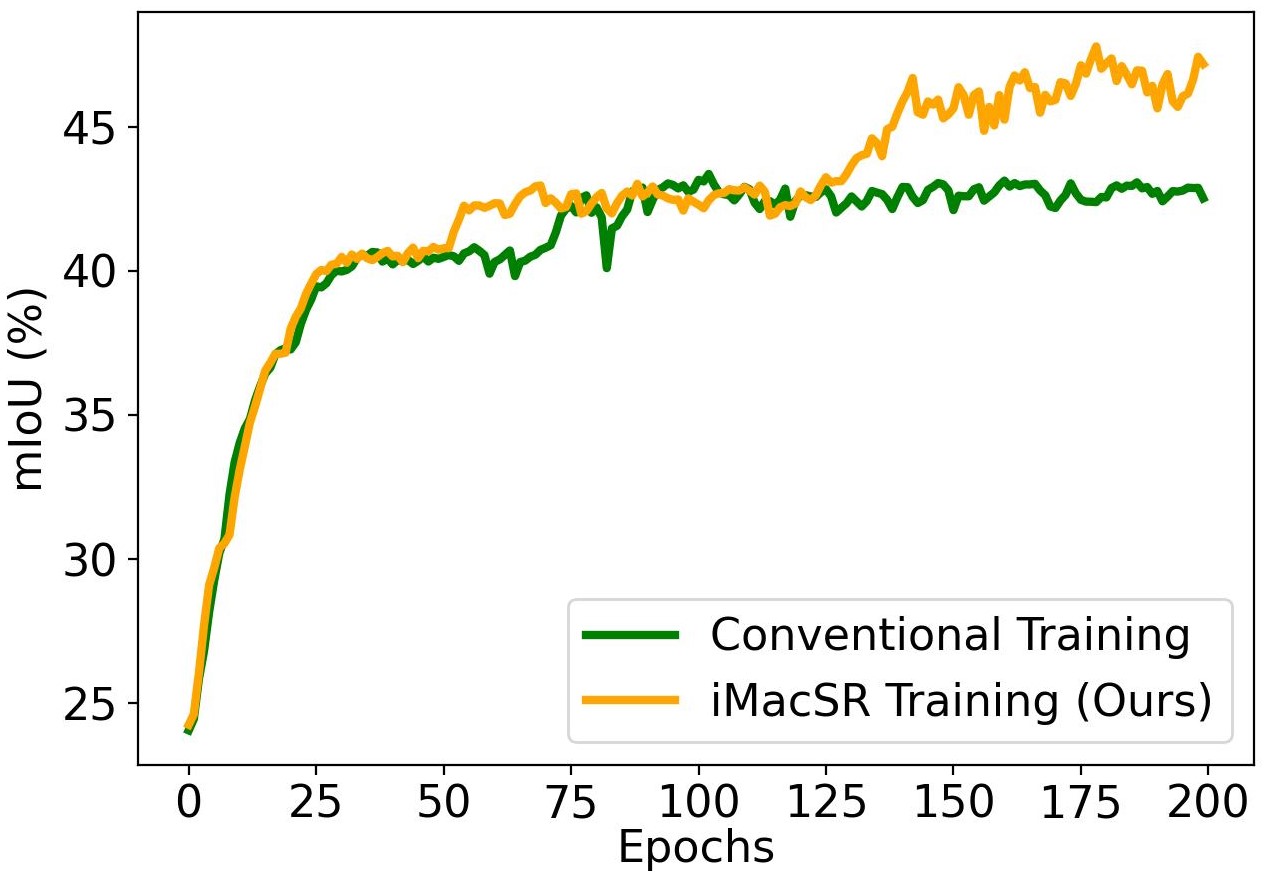}
\label{Fig.iMacRS_Metrics_mIoU}
}
\subfloat[\footnotesize mPrecision]{\includegraphics[width=0.24\linewidth]{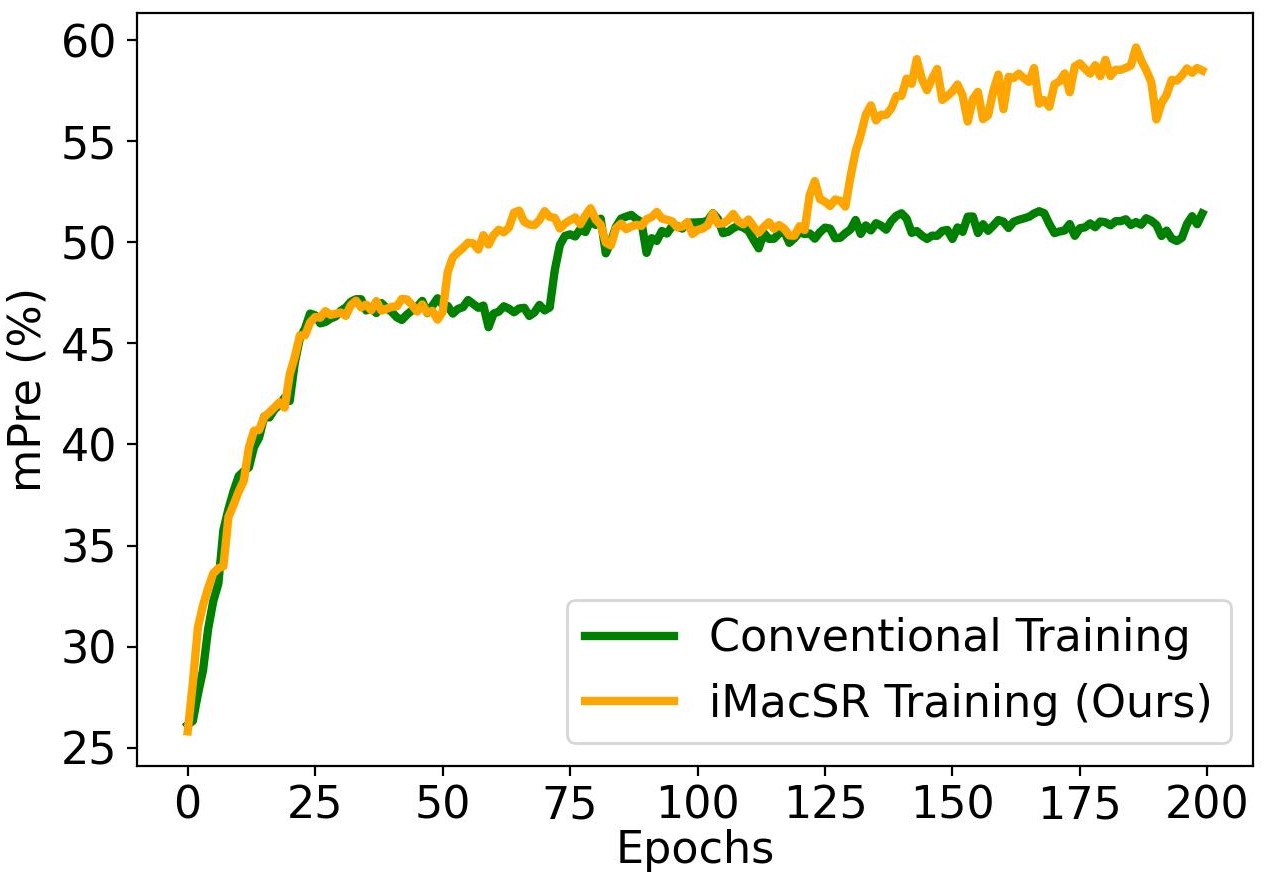}
\label{Fig.iMacRS_Metrics_mPre}
}
\subfloat[\footnotesize mRecall]{\includegraphics[width=0.24\linewidth]{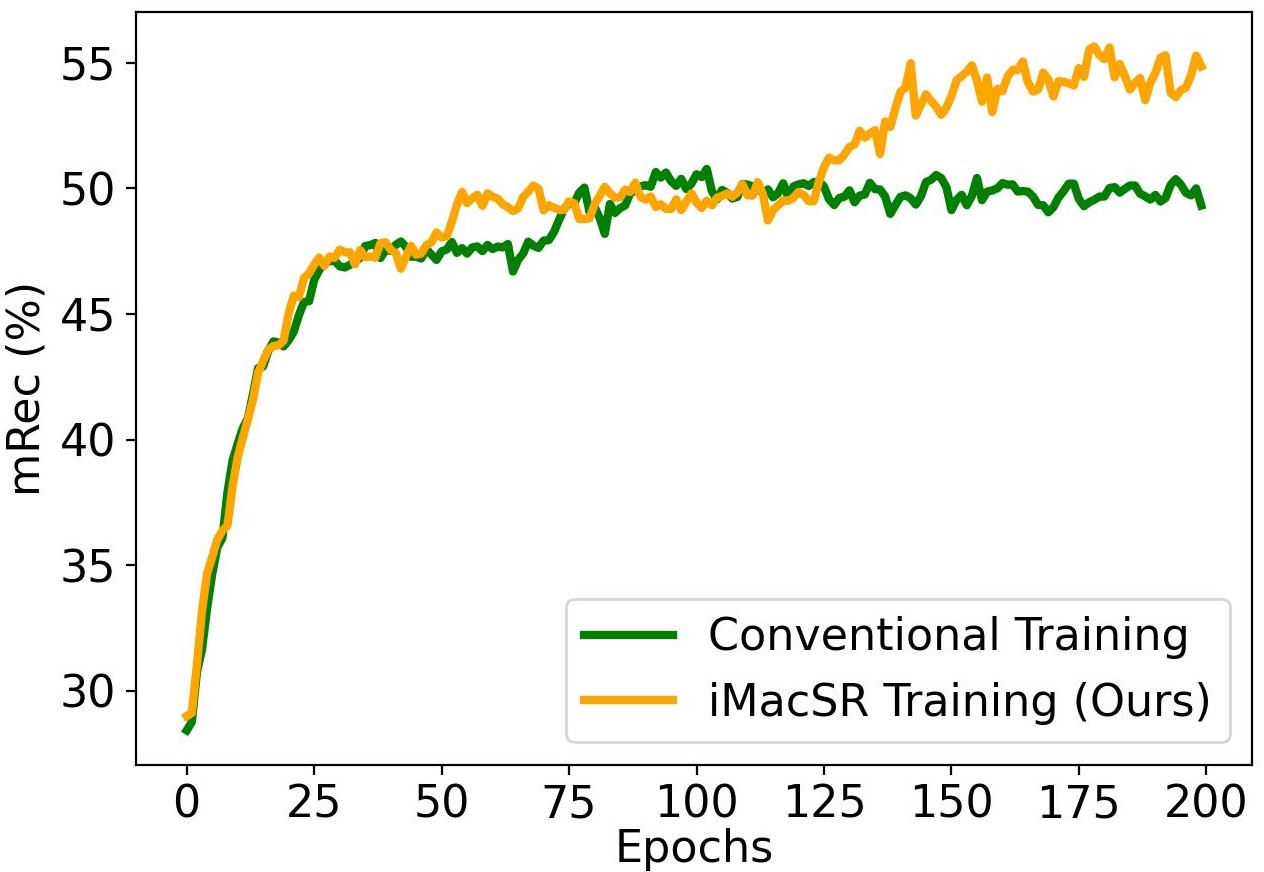}
\label{Fig.iMacRS_Metrics_mRec}
}
\subfloat[\footnotesize mF1]{\includegraphics[width=0.24\linewidth]{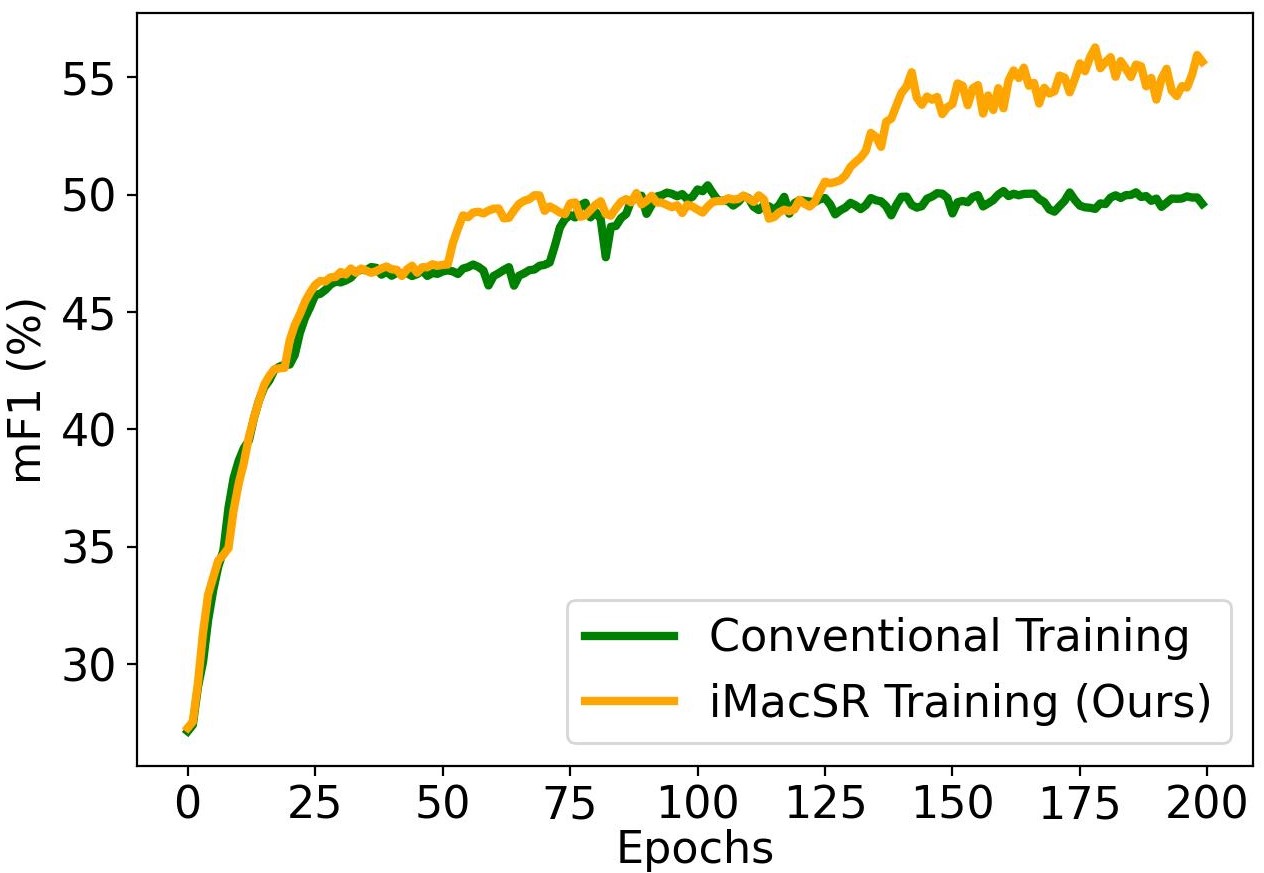}
\label{Fig.iMacRS_Metrics_mF1}
}
\caption{The performance comparison of iMacHSR against the conventional training method for DeepLabv3+ on Cityscapes.}
\label{Fig.iMacRS_quan_Metrics}
\end{figure*}

\subsubsection{Space Complexity.}
Compared to traditional output-layer supervision scheme, the proposed iMacHSR has two extra space consumption parts: latent feature cache and auxiliary dimension adapter. Specifically, the space for latent feature storage is $O(\sum_{m=1}^M Bw_mh_mC_m + M \cdot BWHK)$; the space of auxiliary dimension adapters for $M$ intermediate points is $O(\sum_{m=1}^M P_m)$. In summary, the total extra space is $O(\sum_{m=1}^M\! Bw_mh_mC_m \!+\! MBWHK + \sum_{m=1}^M\! P_m)$.

\subsubsection{Discussion of iMacHSR Complexity.}
Based on above time and space complexity analyses, we can find that iMacHSR introduces linear overheads with respect to $M$ for both time and space. For typical configurations (\eg, $M \leq 5$), this overhead is marginal compared to gain in performance. In practice, choosing $M$ proportional to $\log D$ balances overhead and performance.

\section{Experiments}
\label{experiments}
In this section, we take semantic segmentation task as an example to evaluate the proposed iMacHSR training scheme. These comparisons are based on widely recognized and accepted datasets, model architectures, and metrics.

\subsection{Datasets, Metrics, and Implementation}
\subsubsection{Datasets.}
The Cityscapes dataset \cite{Cordts2016Cityscapes} consists of 2,975 training images and 500 validation images, each annotated with masks. This dataset encompasses 19 semantic classes, such as vehicles and pedestrians. The CamVid dataset \cite{brostow2008segmentation} comprises a total of 701 images across 11 semantic classes. For our experiments, we randomly selected 600 samples for training and used the remaining 101 samples as a test dataset. The SynthiaSF dataset \cite{ros2016synthia} offers a collection of synthetic, yet photorealistic images that emulate urban scenarios. It provides pixel-level annotations for 23 semantic classes, with 1,596 images designated for training and 628 for testing.

\begin{table*}[tp]
\centering
\renewcommand{\arraystretch}{0.24}
\addtolength{\tabcolsep}{-0.52pt}
\begin{tabularx}{\linewidth}{|l|lllll|}
\hline
\verticaltext[27.5pt]{Raw RGBs} &
\includegraphics[width=0.187\linewidth, height=0.12\linewidth]{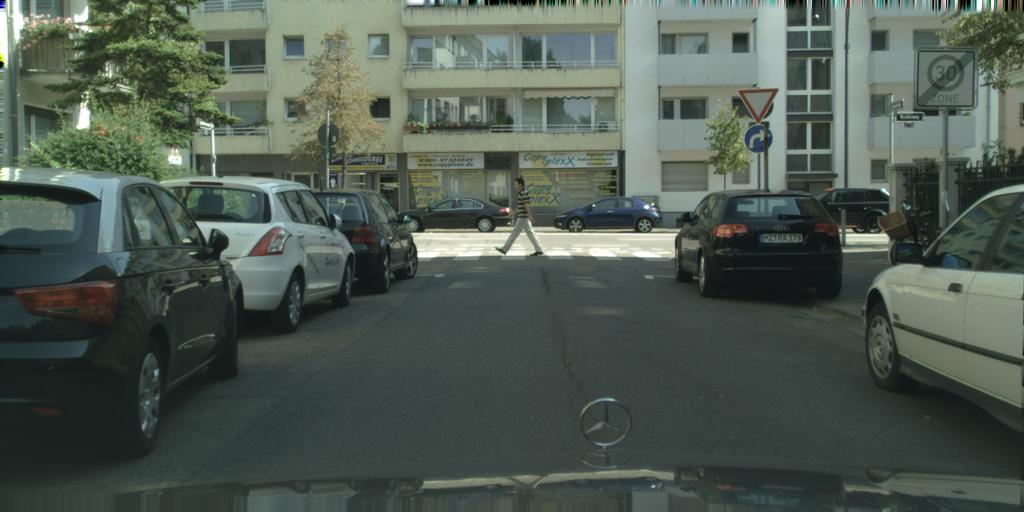} &\hspace{-0.47cm}
\includegraphics[width=0.187\linewidth, height=0.12\linewidth]{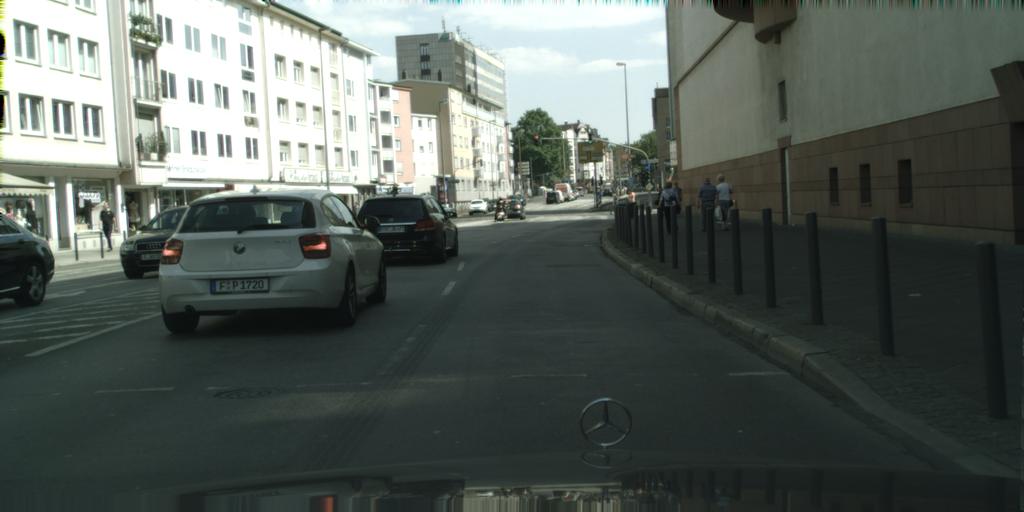} &\hspace{-0.47cm}
\includegraphics[width=0.187\linewidth, height=0.12\linewidth]{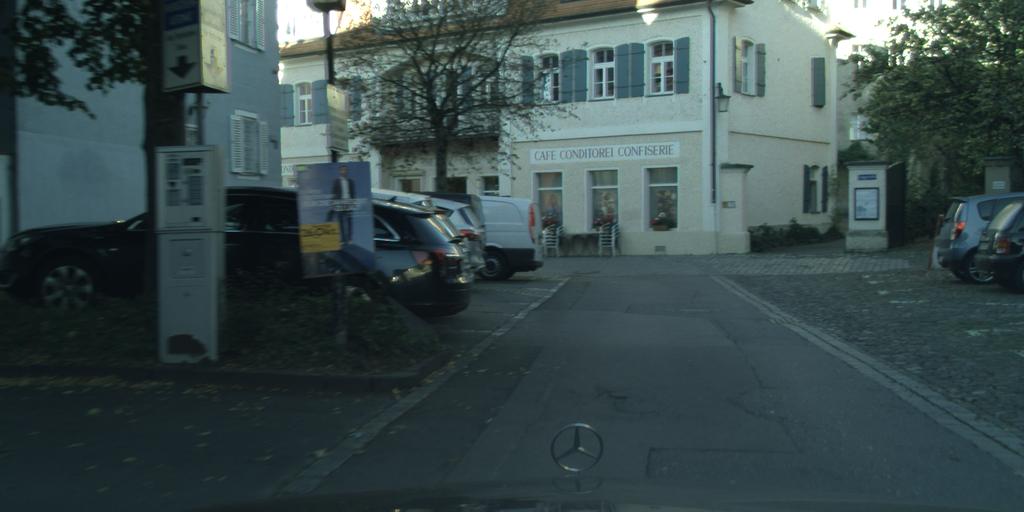   } &\hspace{-0.47cm}
\includegraphics[width=0.187\linewidth, height=0.12\linewidth]{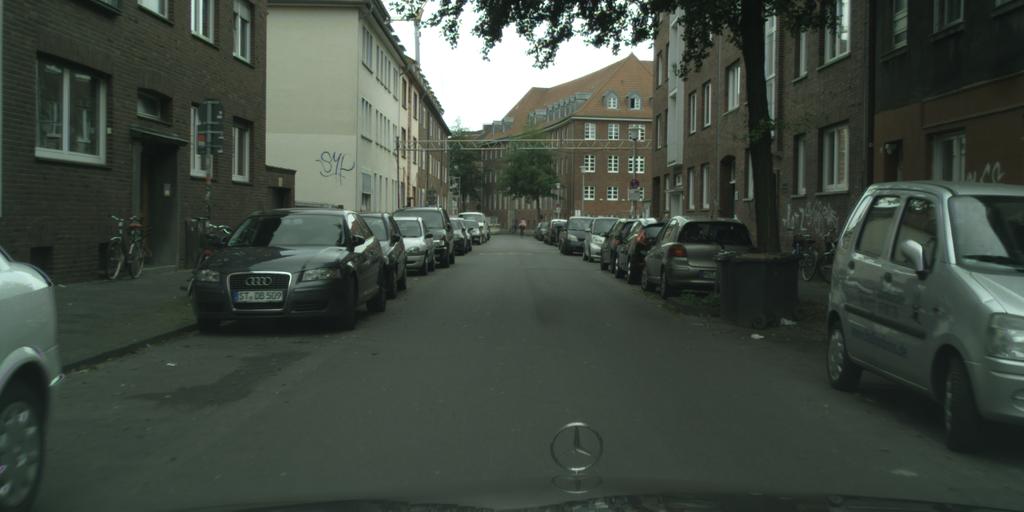  } &\hspace{-0.47cm}
\includegraphics[width=0.187\linewidth, height=0.12\linewidth]{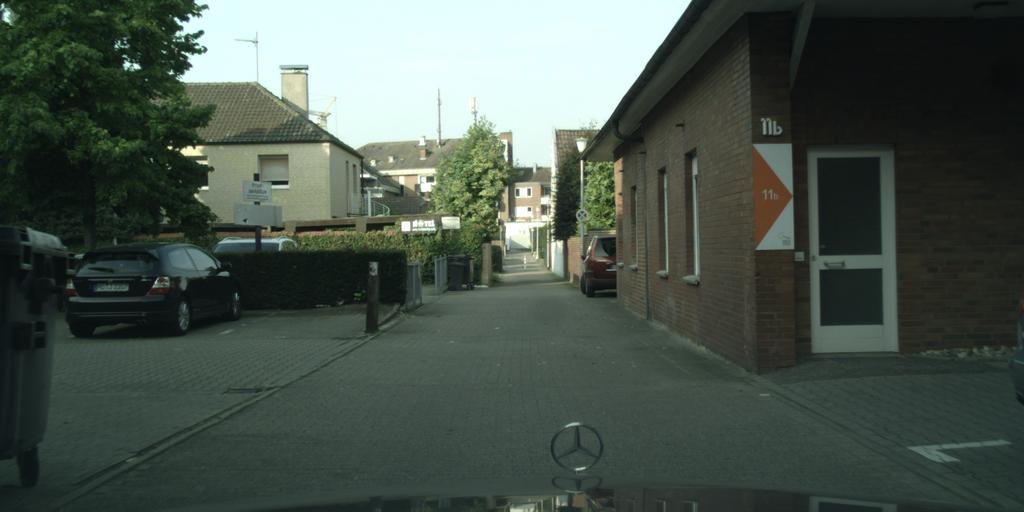  }\\
\hline

\verticaltext[27.5pt]{Ground Truth} &
\includegraphics[width=0.187\linewidth, height=0.12\linewidth]{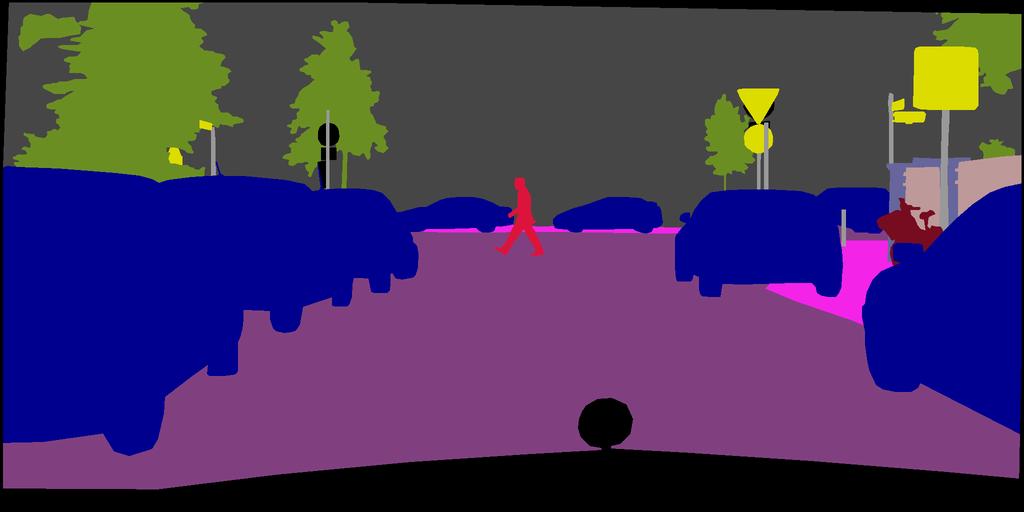} &\hspace{-0.47cm}
\includegraphics[width=0.187\linewidth, height=0.12\linewidth]{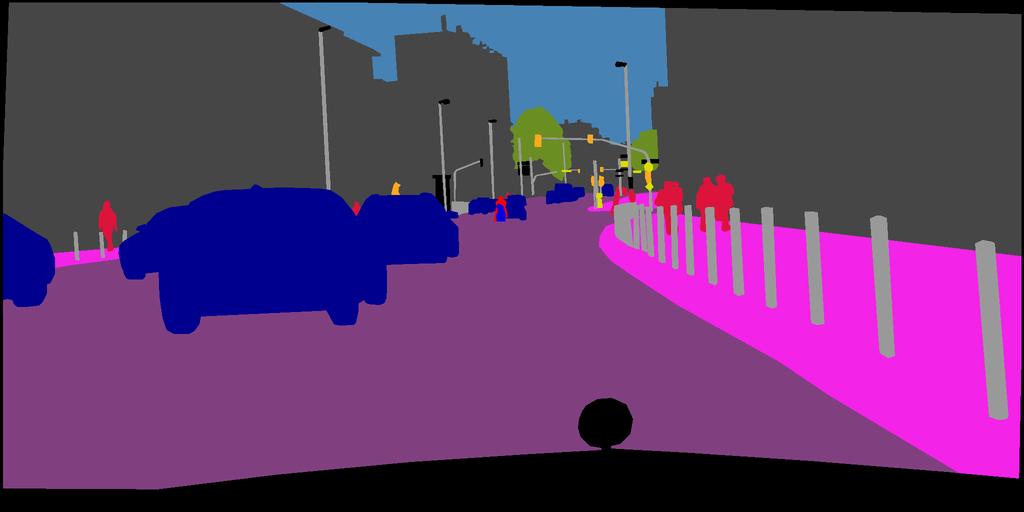} &\hspace{-0.47cm}
\includegraphics[width=0.187\linewidth, height=0.12\linewidth]{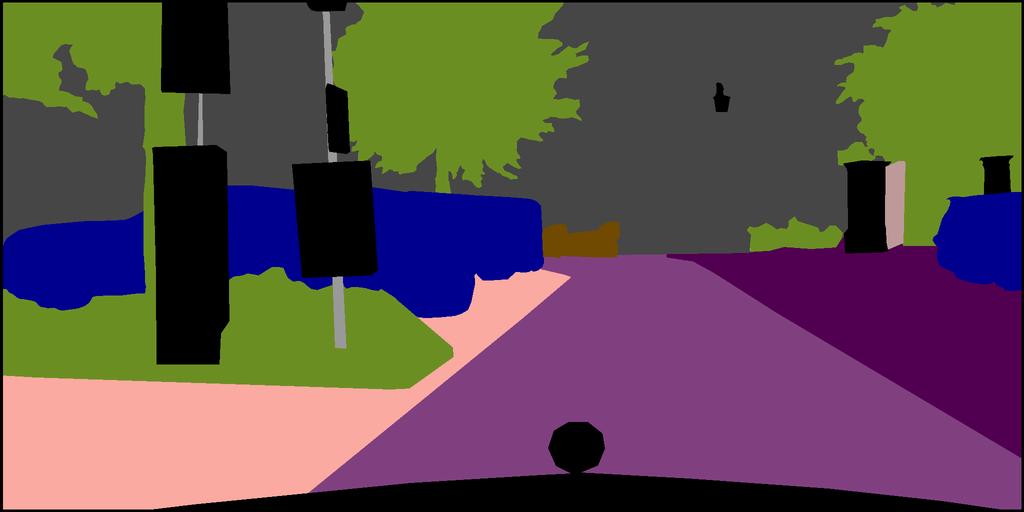   } &\hspace{-0.47cm}
\includegraphics[width=0.187\linewidth, height=0.12\linewidth]{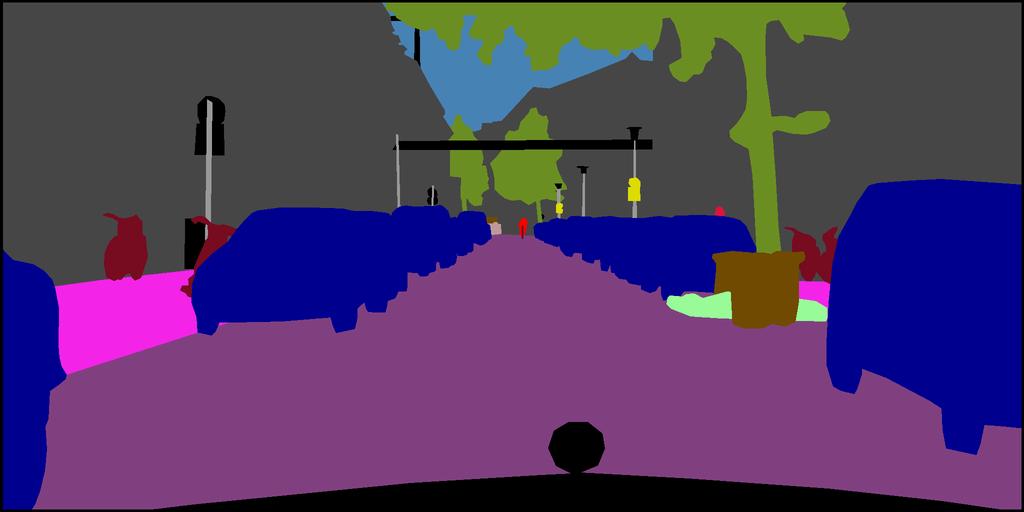  } &\hspace{-0.47cm}
\includegraphics[width=0.187\linewidth, height=0.12\linewidth]{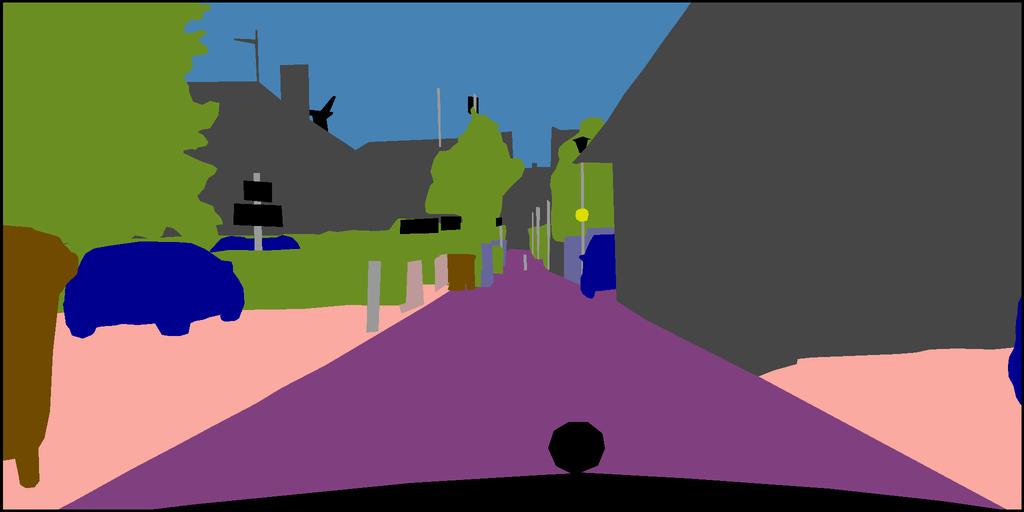  }\\
\hline

\verticaltext[27.5pt]{Dis. iMacHSR} &
\includegraphics[width=0.187\linewidth, height=0.12\linewidth]{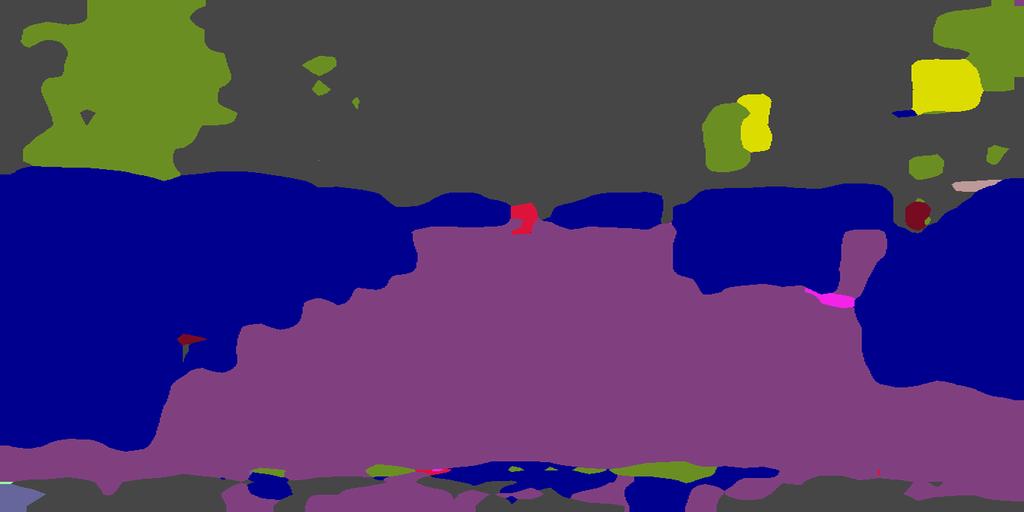} &\hspace{-0.47cm}
\includegraphics[width=0.187\linewidth, height=0.12\linewidth]{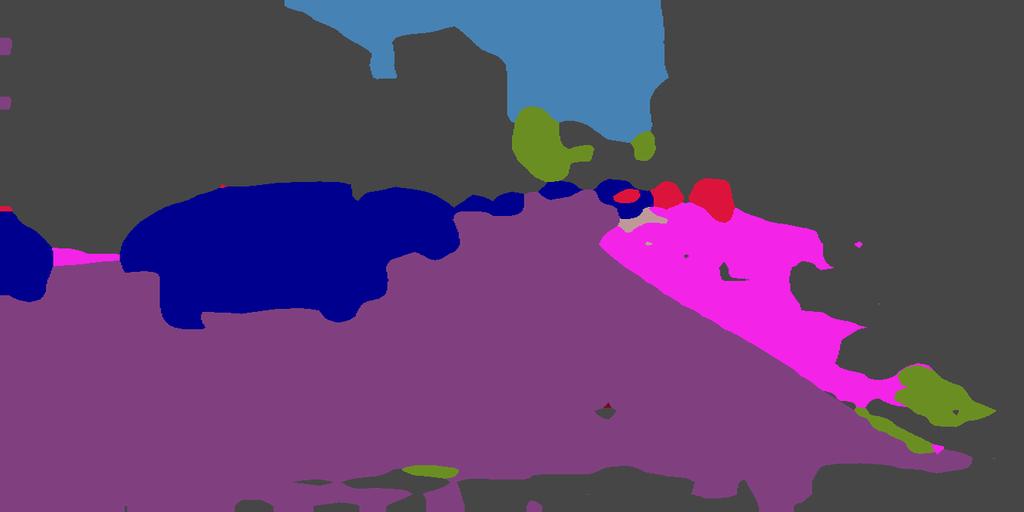} &\hspace{-0.47cm}
\includegraphics[width=0.187\linewidth, height=0.12\linewidth]{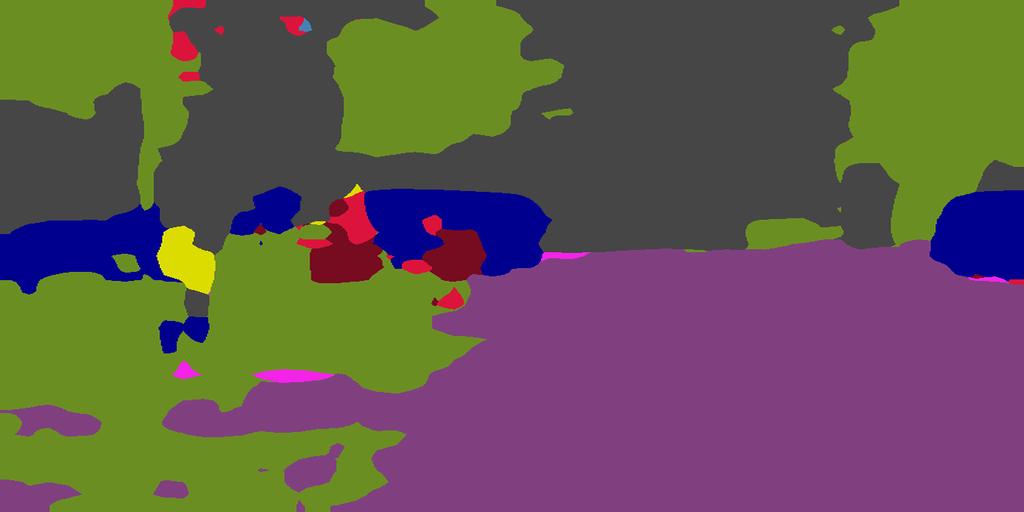   } &\hspace{-0.47cm}
\includegraphics[width=0.187\linewidth, height=0.12\linewidth]{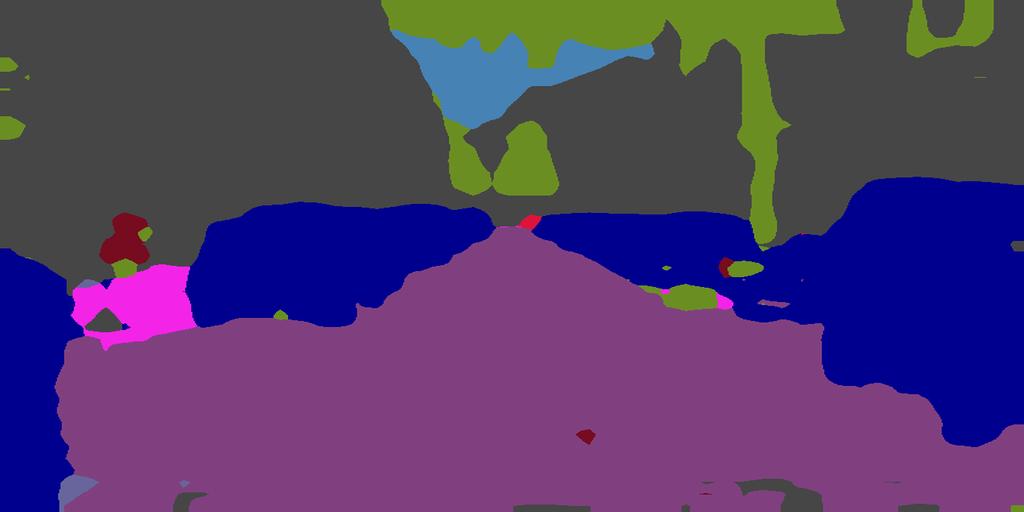  } &\hspace{-0.47cm}
\includegraphics[width=0.187\linewidth, height=0.12\linewidth]{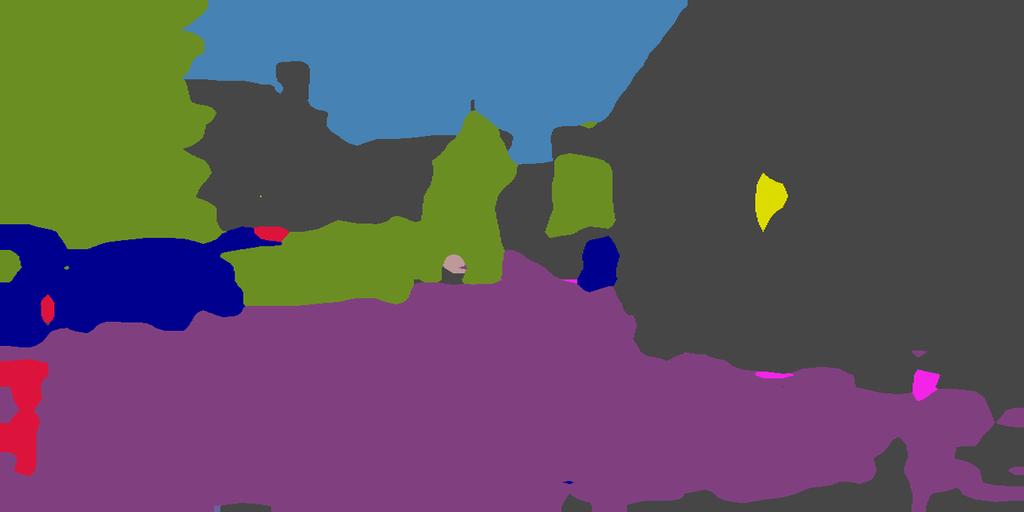  }\\
\hline

\verticaltext[27.5pt]{\textbf{En. iMacHSR}} &
\includegraphics[width=0.187\linewidth, height=0.12\linewidth]{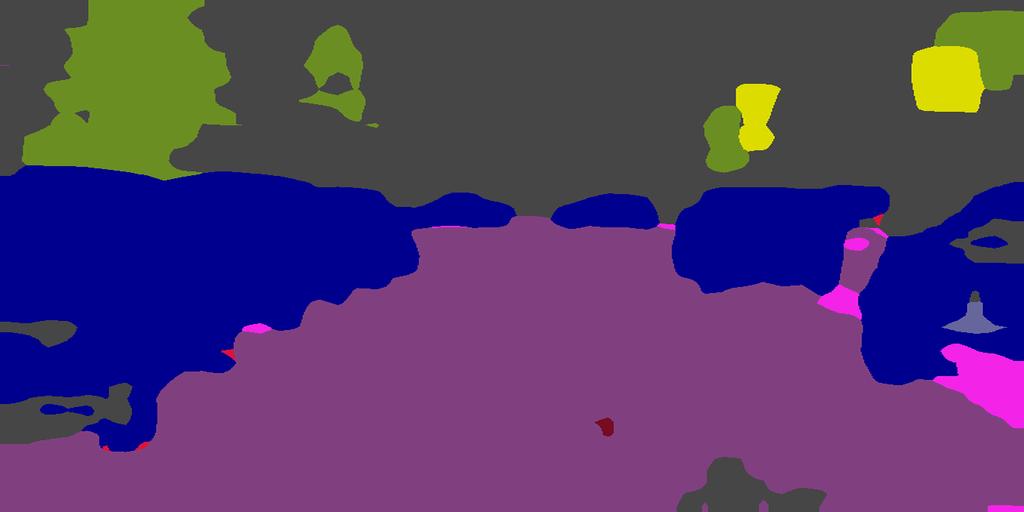} &\hspace{-0.47cm}
\includegraphics[width=0.187\linewidth, height=0.12\linewidth]{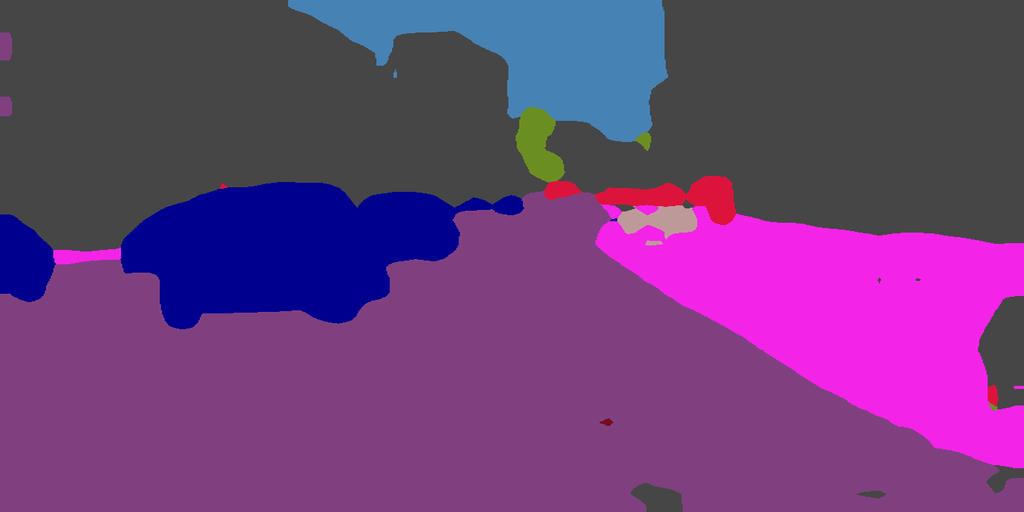} &\hspace{-0.47cm}
\includegraphics[width=0.187\linewidth, height=0.12\linewidth]{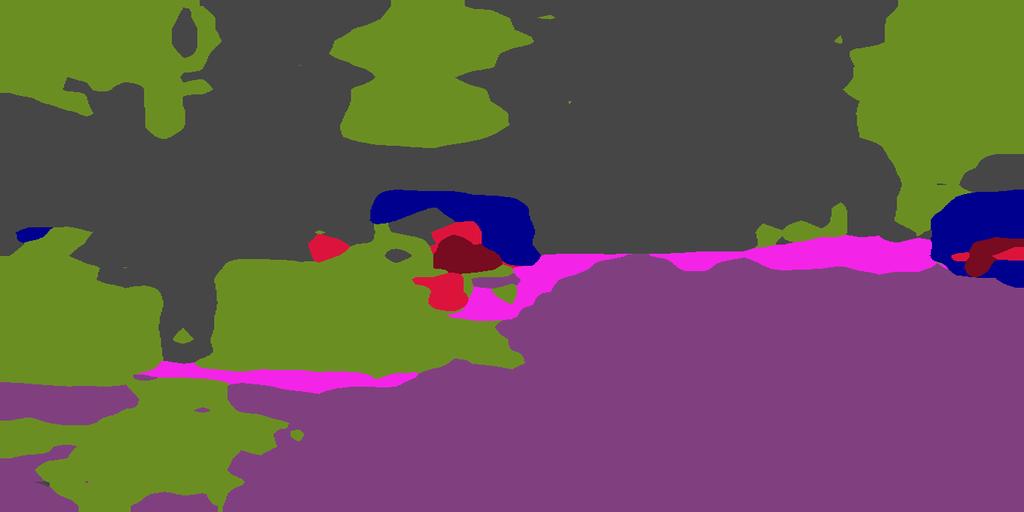   } &\hspace{-0.47cm}
\includegraphics[width=0.187\linewidth, height=0.12\linewidth]{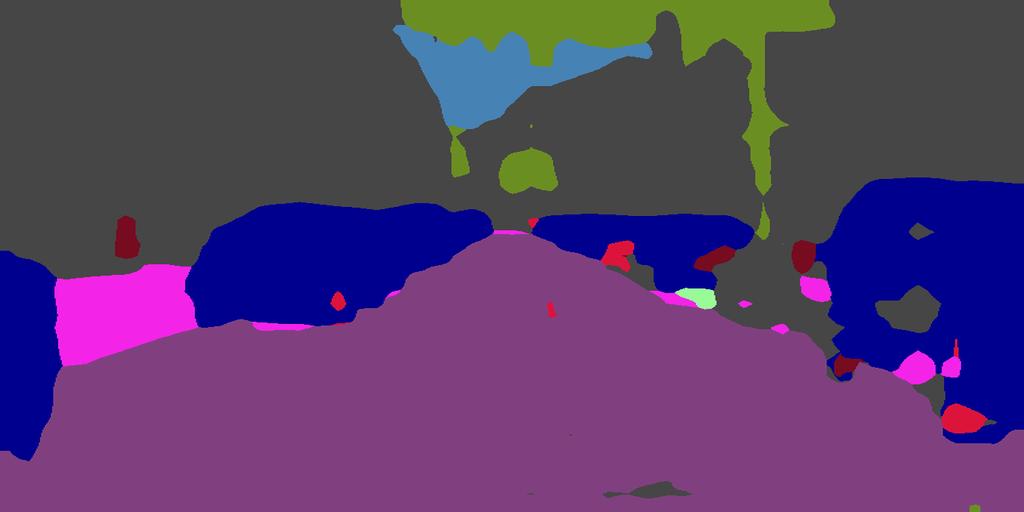  } &\hspace{-0.47cm}
\includegraphics[width=0.187\linewidth, height=0.12\linewidth]{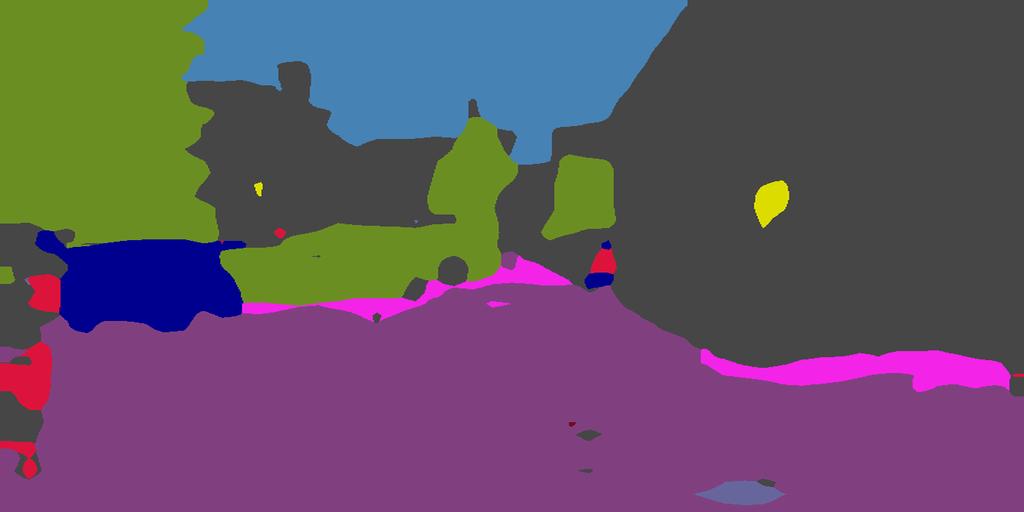  }\\
\hline
\end{tabularx}
\caption{Qualitative performance comparison of the proposed iMacHSR against conventional training method}
\label{tab:iMacRS_qualitive_comp}
\end{table*}

\subsubsection{Evaluation Metrics.}
We assess the proposed iMacHSR on semantic segmentation task using four commonly used metrics: mIoU, mPrecision (mPre for short), mRecall (mRec for short), and mF1. These metrics are formulated in \textit{Appendix III of Supplementary Materials}.

\subsubsection{Implementation Details.}
The primary configurations of hardware, software, and the detailed training parameters are outlined in \textit{Appendix IV of Supplementary Materials}. Our experiments include a comparative analysis of the proposed iMacHSR training method against traditional output-layer supervision taining approach across three models—DeepLabv3+ \cite{chen2018encoderdecoder}, TopFormer \cite{zhang2022topformer}, and SeaFormer \cite{wan2023seaformer}—on three datasets, namely Cityscapes, CamVid, and SynthiaSF.

\subsection{Main Results and Empirical Analyses}
\subsubsection{Quantitative Performance Comparison.}

\begin{figure}[t]
\centering
\subfloat[\footnotesize mIoU]{\includegraphics[width=0.48\linewidth]{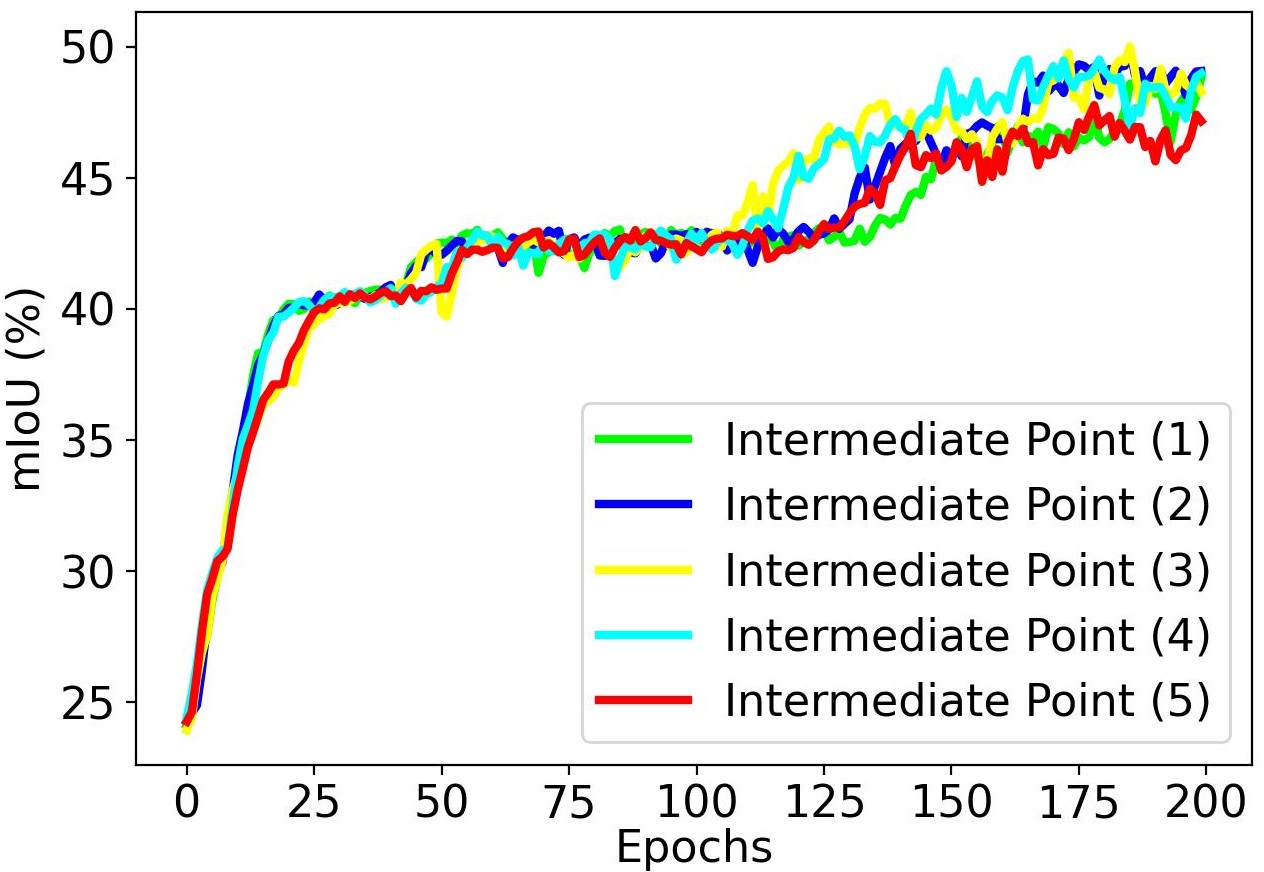}
\label{Fig.abl_number_Metrics_mIoU}
}
\subfloat[\footnotesize mF1]{\includegraphics[width=0.48\linewidth]{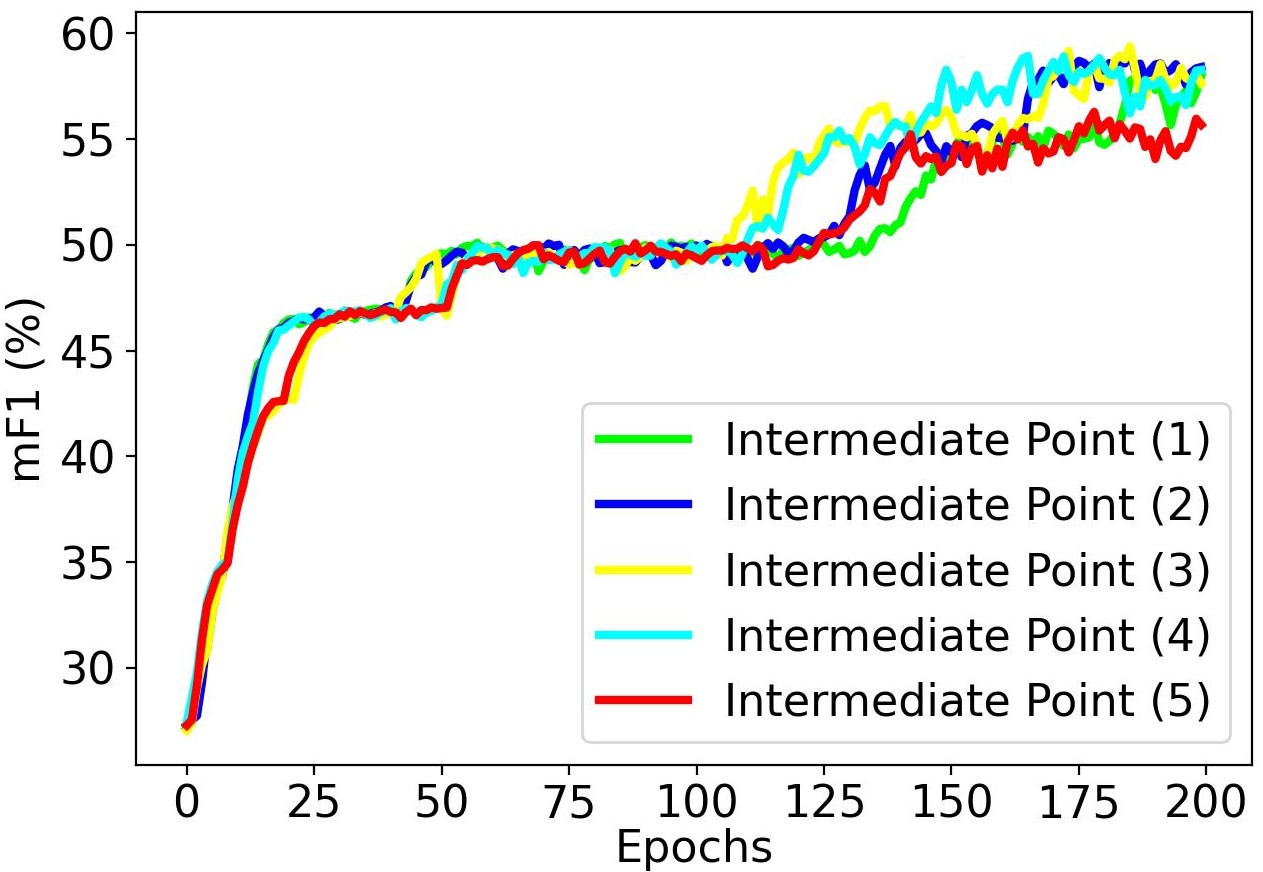}
\label{Fig.abl_number_Metrics_mF1}
}
\caption{The impact of the number of intermediate points on iMacHSR's training performance.}
\label{Fig.abl_number_Metrics}
\end{figure}

\begin{figure}[t]
\centering
\subfloat[\footnotesize mIoU]{\includegraphics[width=0.48\linewidth]{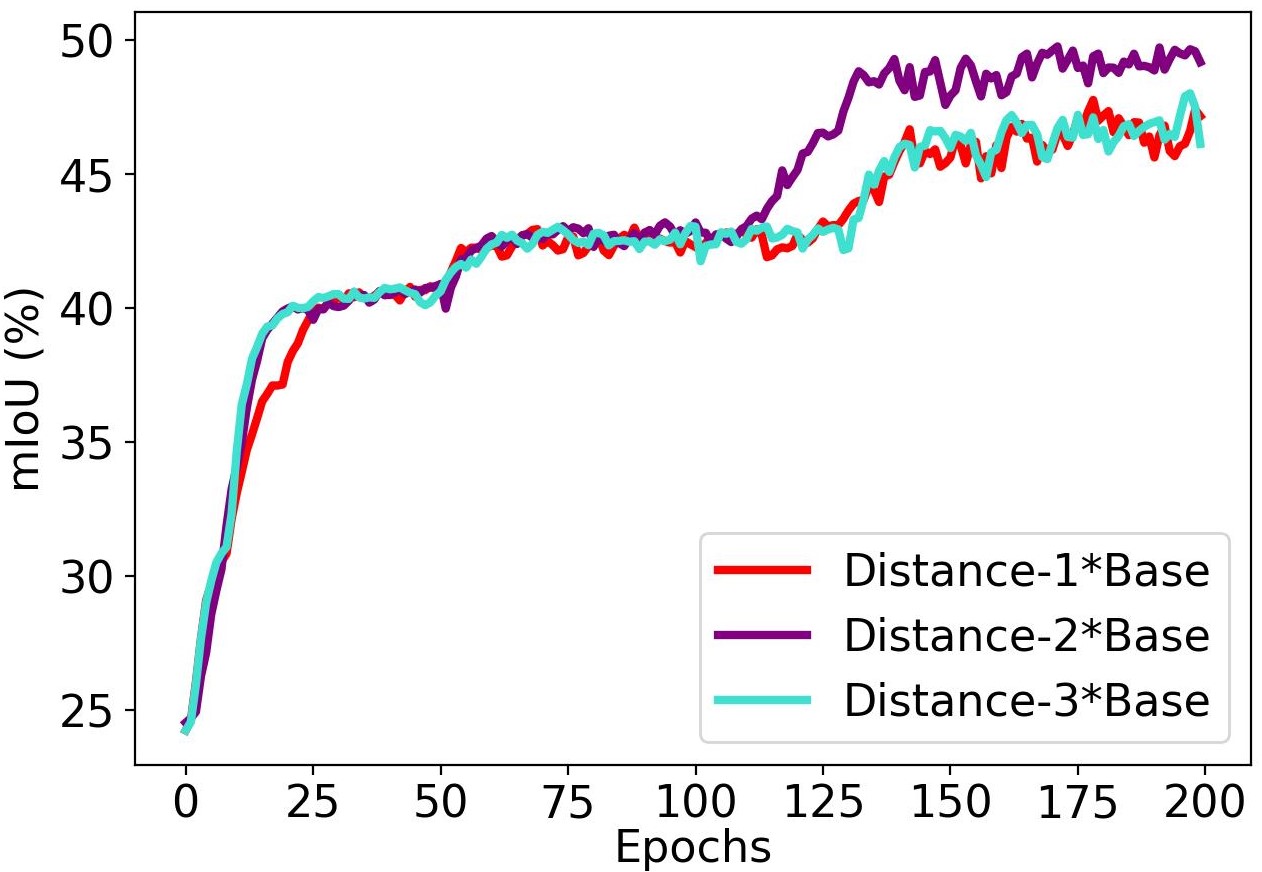}
\label{Fig.abl_dist_Metrics_mIoU}
}
\subfloat[\footnotesize mF1]{\includegraphics[width=0.48\linewidth]{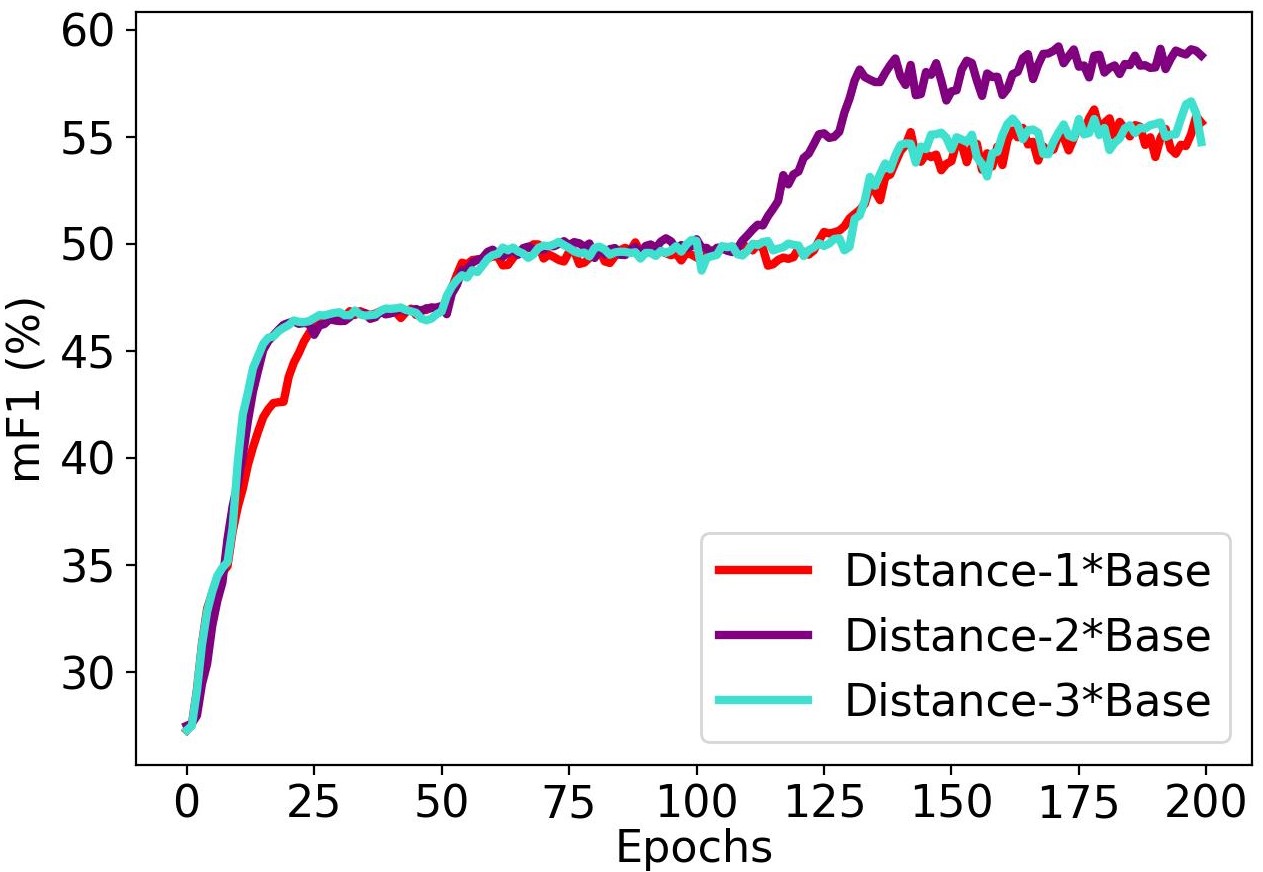}
\label{Fig.abl_dist_Metrics_mF1}
}
\caption{The impact of the distance between adjacent intermediate points on iMacHSR's training performance.}
\label{Fig.abl_dist_Metrics}
\end{figure}

We carry out a bunch of experiments to compare the quantitative performance of enabling the proposed iMacHSR training scheme against disabling iMacHSR training scheme on CNN-based DeepLabv3+ model, and Transformer-based SeaFormer and TopFormer models. The results for all adopted models are presented in \Cref{tab:iMacRS_quantitative_comp}. From \Cref{tab:iMacRS_quantitative_comp}, we can conclude following insights: (I) The case of enabling iMacHSR exceeds the case of disabling iMacHSR in performance for all adopted models across almost all metrics on Cityscapes, CamVid, and SynthiaSF datasets. This effectively demonstrates the superiority of the proposed iMacHSR. Taking the combination of DeepLabv3+ model and Cityscapes dataset as an example, the case of enabling iMacHSR outperforms the case of disabling iMacHSR by margins of (47.78 - 43.76) / 43.76 \%  = 9.19\%, (56.28 - 50.40) / 50.40 \%  = 11.67\%, (59.64 - 51.54) / 51.54 \%  = 15.72\%, and (55.64 - 50.77) / 50.77 \%  = 9.59\% in mIoU, mF1, mPrecision, mRecall, respectively. This great enhancement in performance can be further visually confirmed in \Cref{Fig.iMacRS_quan_Metrics}. (II) The performance improvement of enabling iMacHSR relative to disabling iMacHSR sometimes is related to the complexity of dataset. Specifically, the more complex the dataset, the greater the performance is enhanced. For example, iMacHSR improves DeepLabv3+ performance in mIoU by 9.19\% on Cityscapes dataset, while by (76.13 - 76.02) / 76.02 \%  = 0.14\% and (34.28 - 33.28) / 33.28 \%  = 3.00\% on CamVid dataset and SythiaSF dataset, respectively. (III) The model architecture sometimes also impacts the performance improvement of the proposed iMacHSR. For example, on SynthiaSF dataset, the proposed iMacHSR can improve the performance of DeepLabv3+ model and TopFormer model, but it fails to improve the performance of SeaFormer model.

\begin{table*}[tp]
\centering
\renewcommand{\arraystretch}{0.40}
\addtolength{\tabcolsep}{-0.4pt}
\begin{tabularx}{\linewidth}{|c|cccc|}
\hline
& \textbf{mIoU} & \textbf{mPre} & \textbf{mRec } & \textbf{mF1} \\
\hline

\verticaltext[38pt]{\textbf{Single-Point}} &\hspace{-0.21cm}
\includegraphics[width=0.235\linewidth]{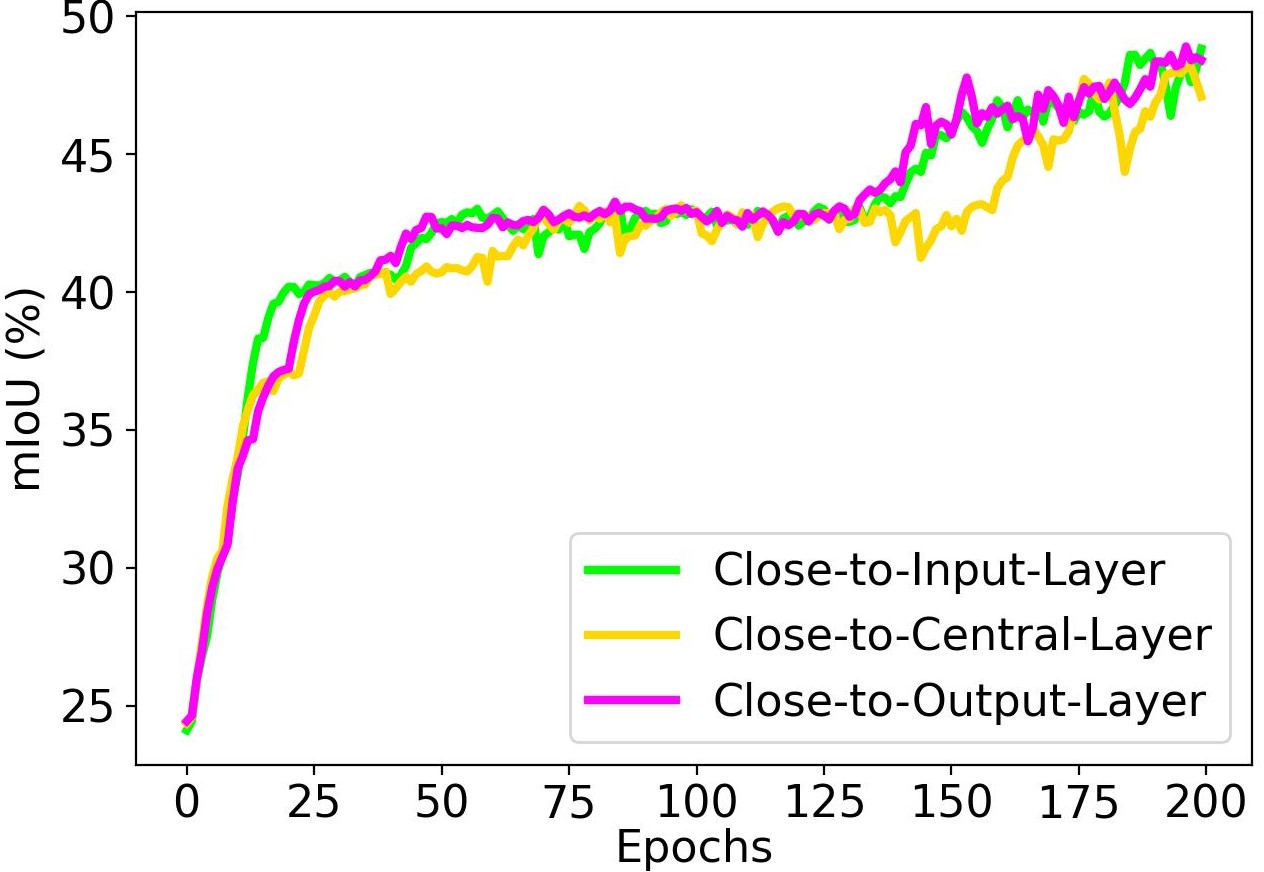} &\hspace{-0.47cm}
\includegraphics[width=0.235\linewidth]{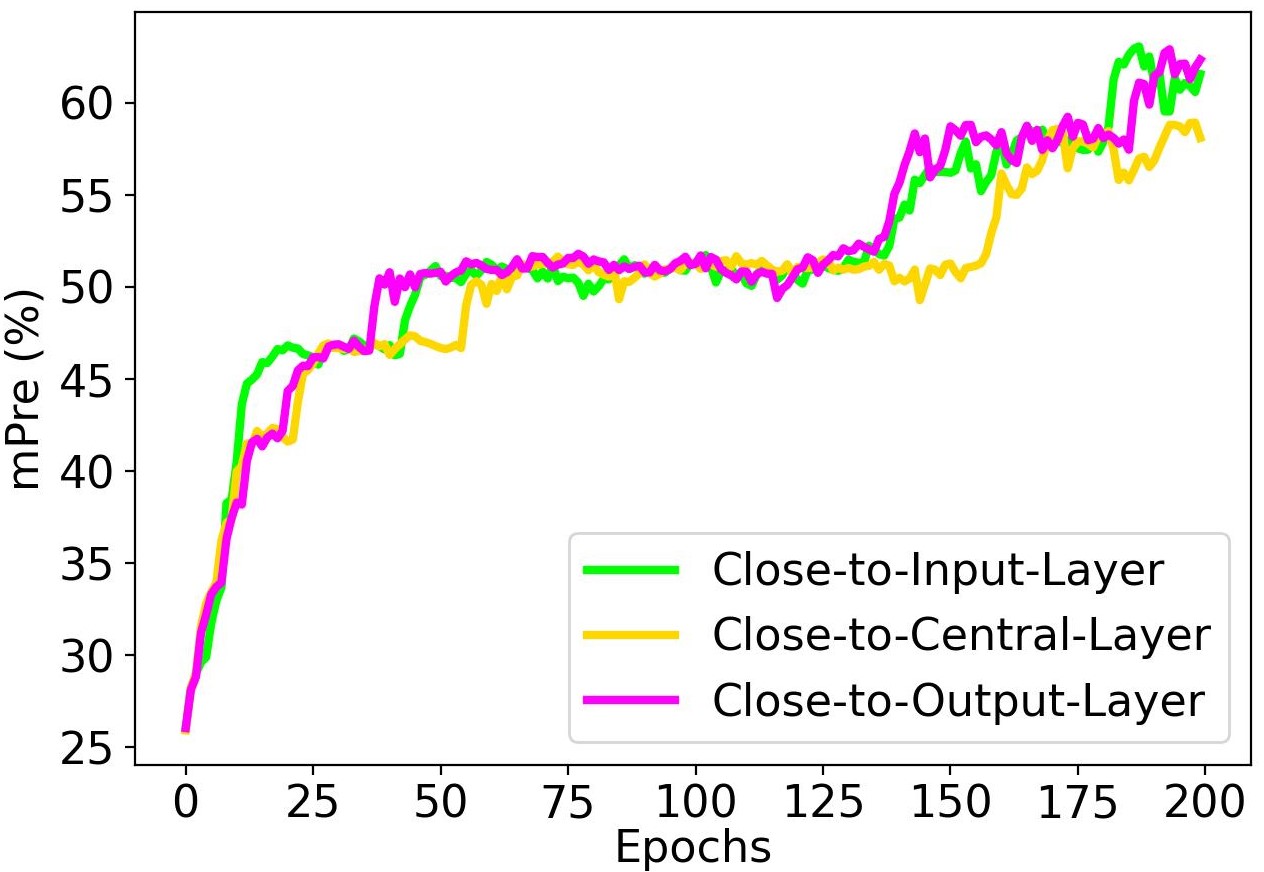} &\hspace{-0.47cm}
\includegraphics[width=0.235\linewidth]{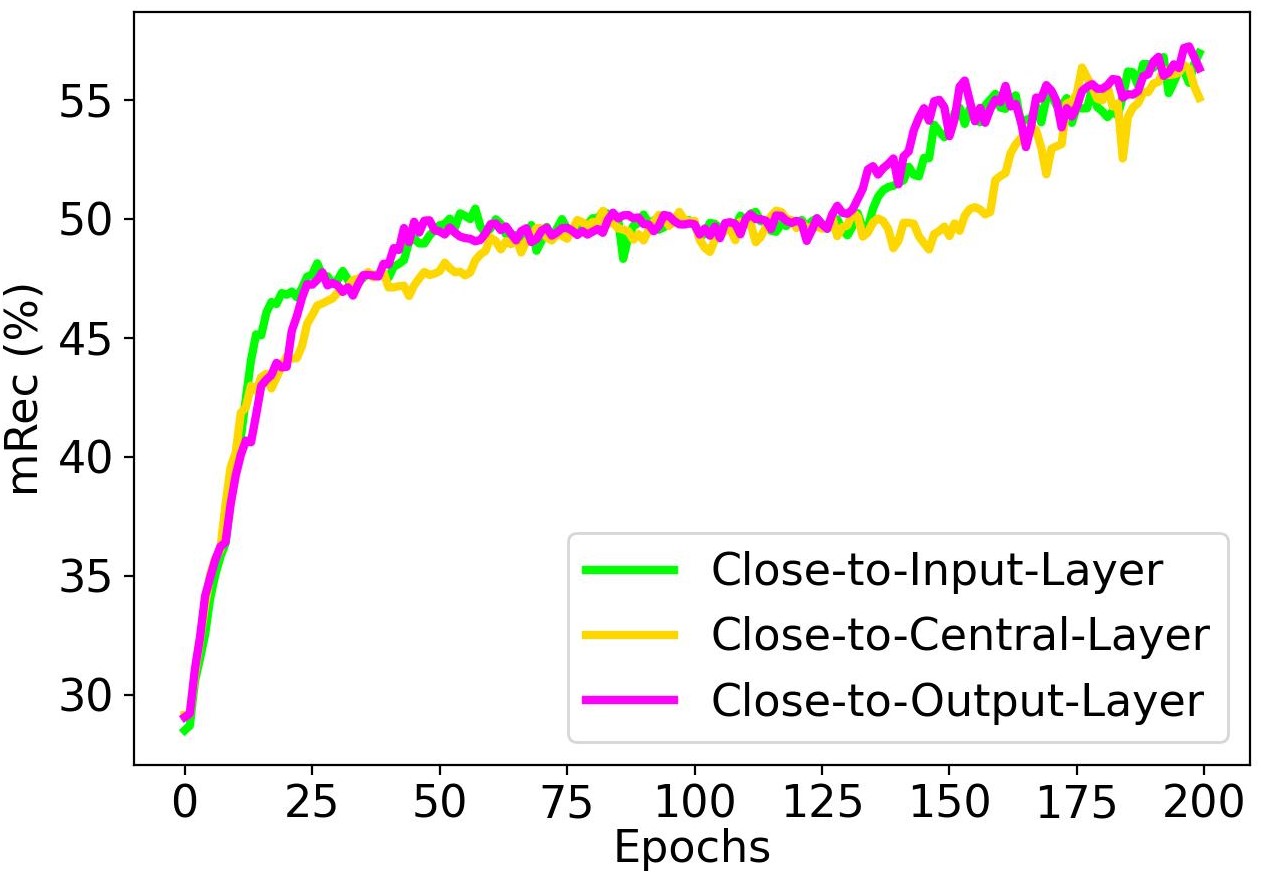} &\hspace{-0.47cm}
\includegraphics[width=0.235\linewidth]{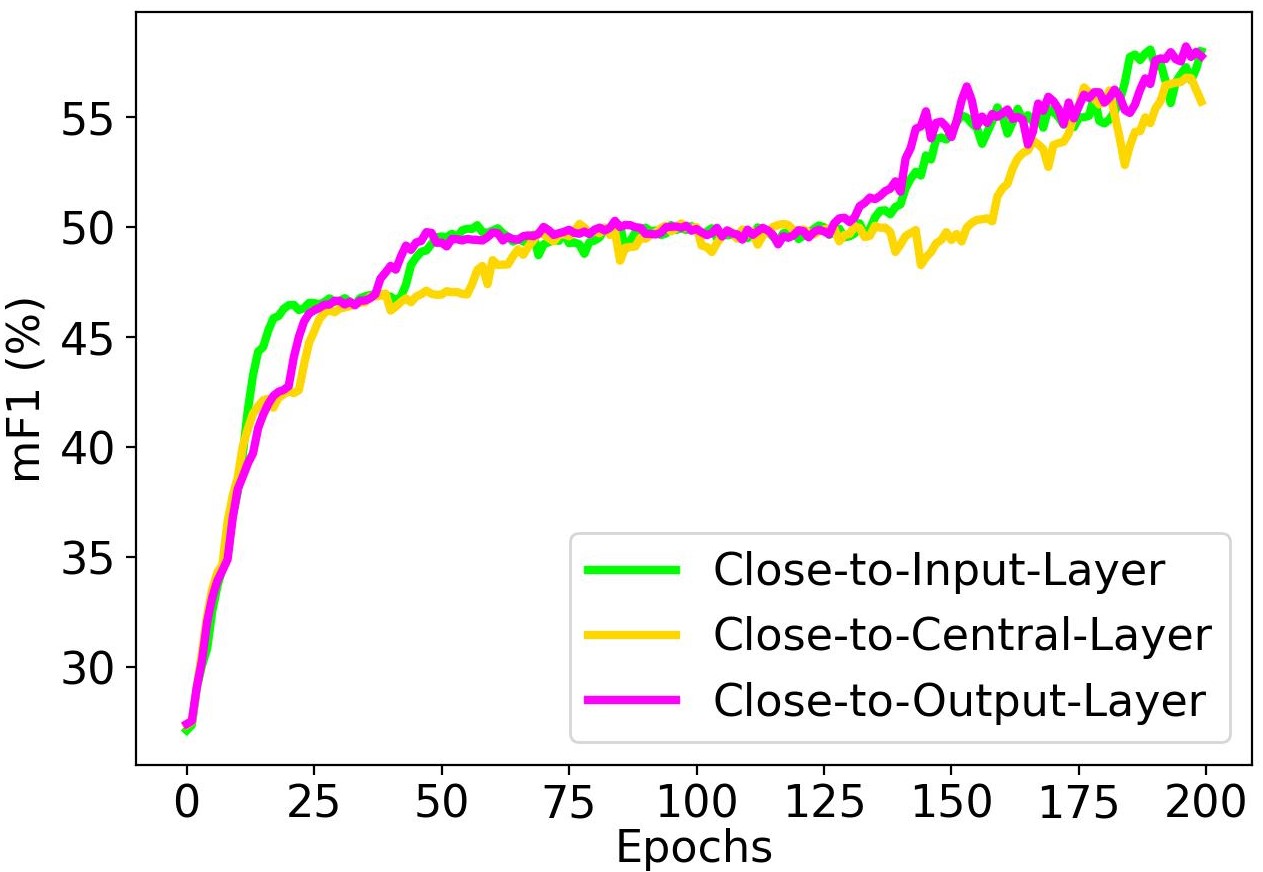 } \\
\hline

\verticaltext[38pt]{\textbf{Two-Point}} &\hspace{-0.21cm}
\includegraphics[width=0.235\linewidth]{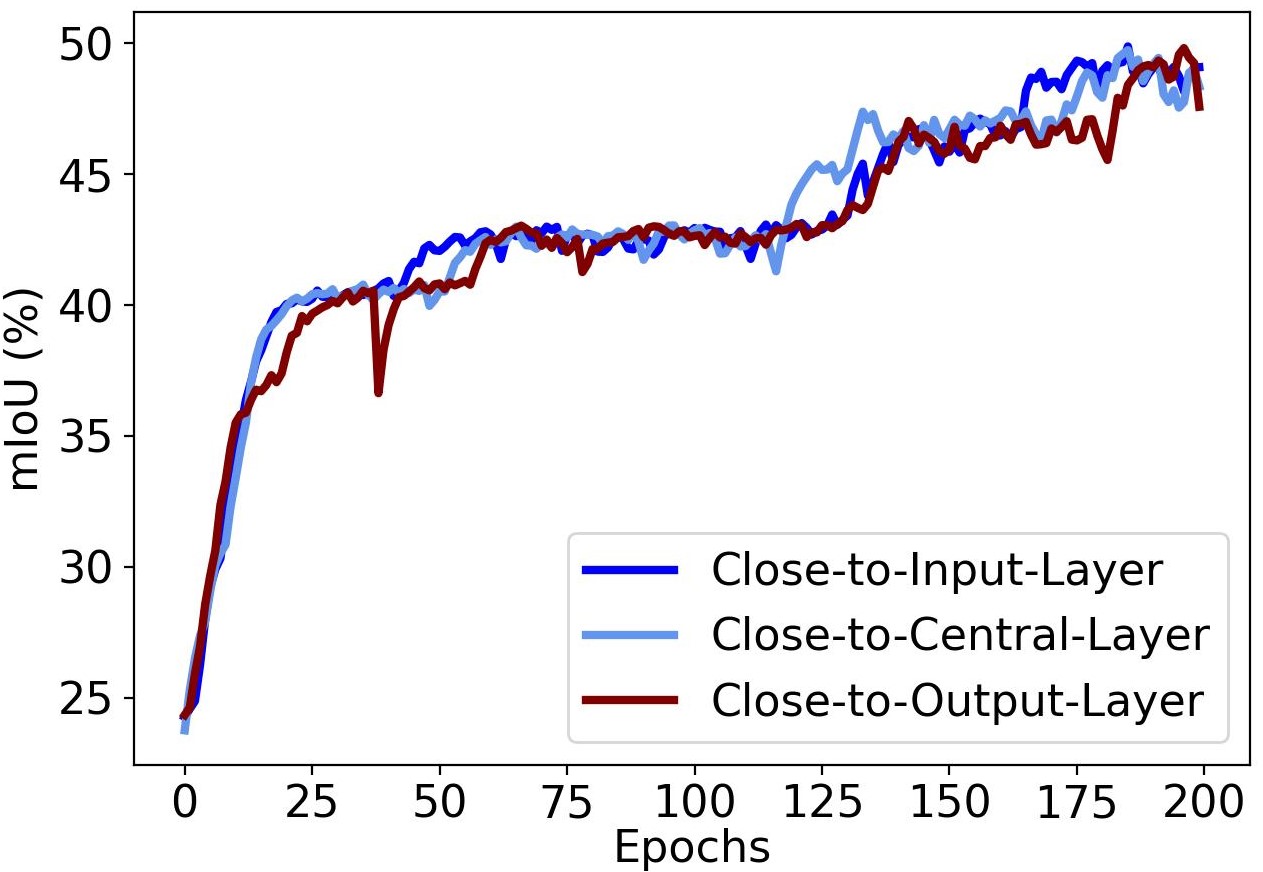} &\hspace{-0.47cm}
\includegraphics[width=0.235\linewidth]{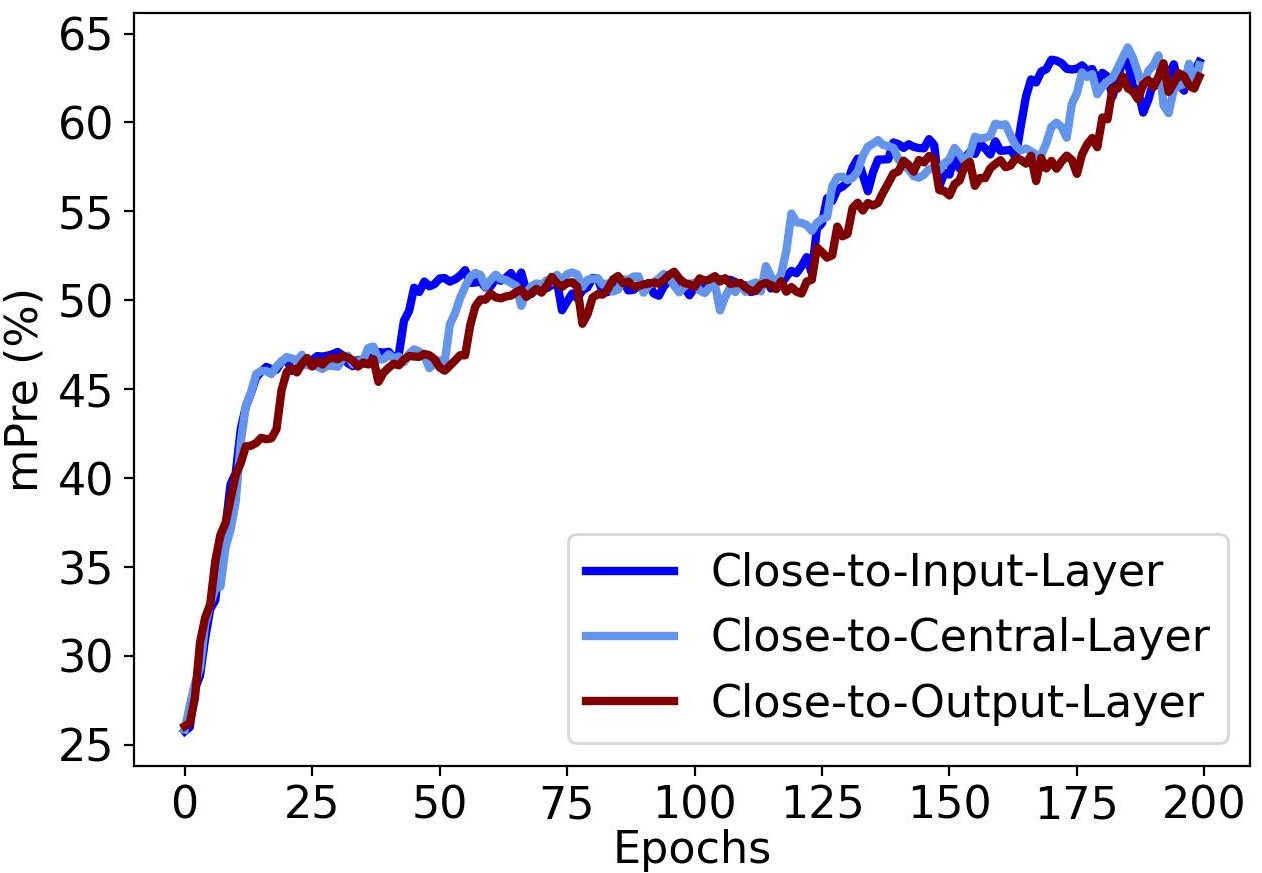} &\hspace{-0.47cm}
\includegraphics[width=0.235\linewidth]{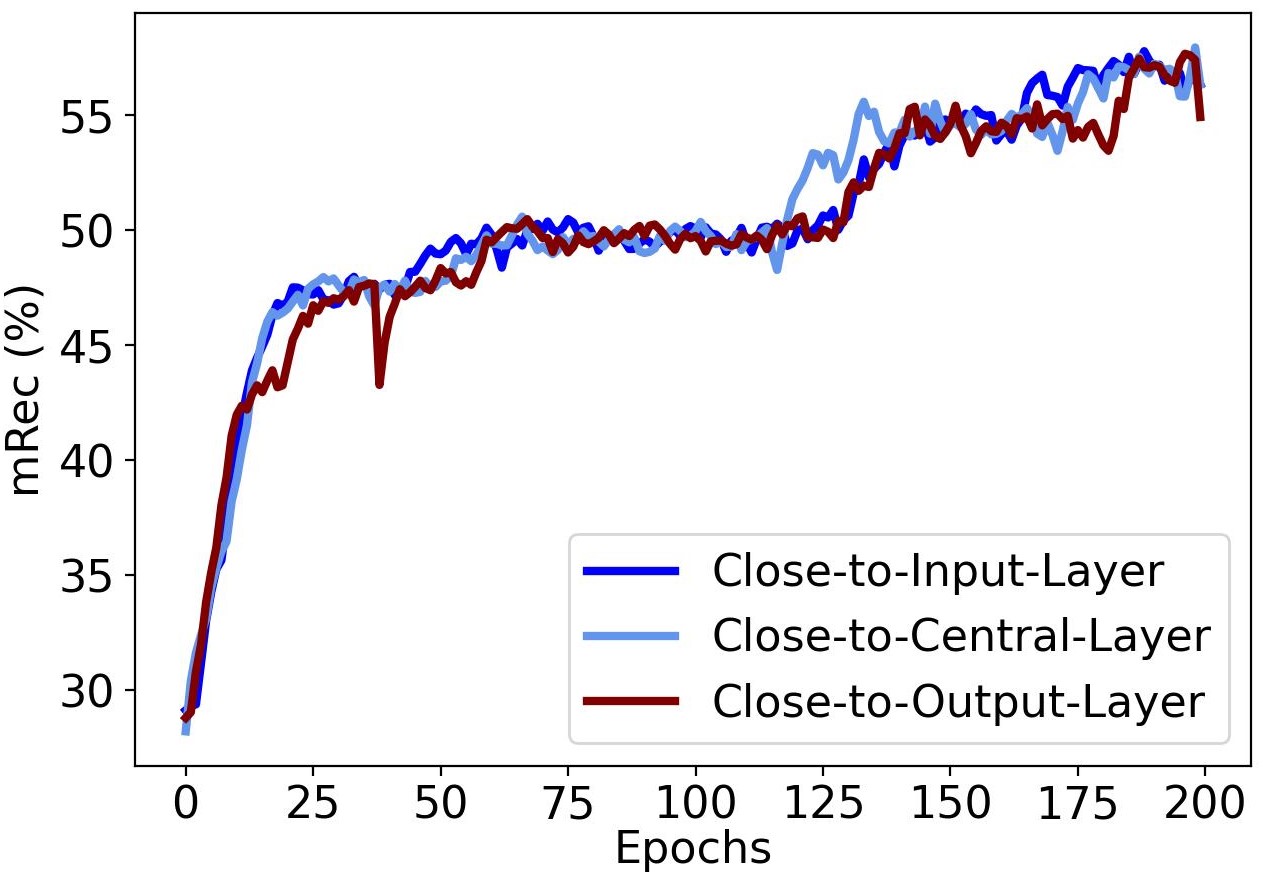} &\hspace{-0.47cm}
\includegraphics[width=0.235\linewidth]{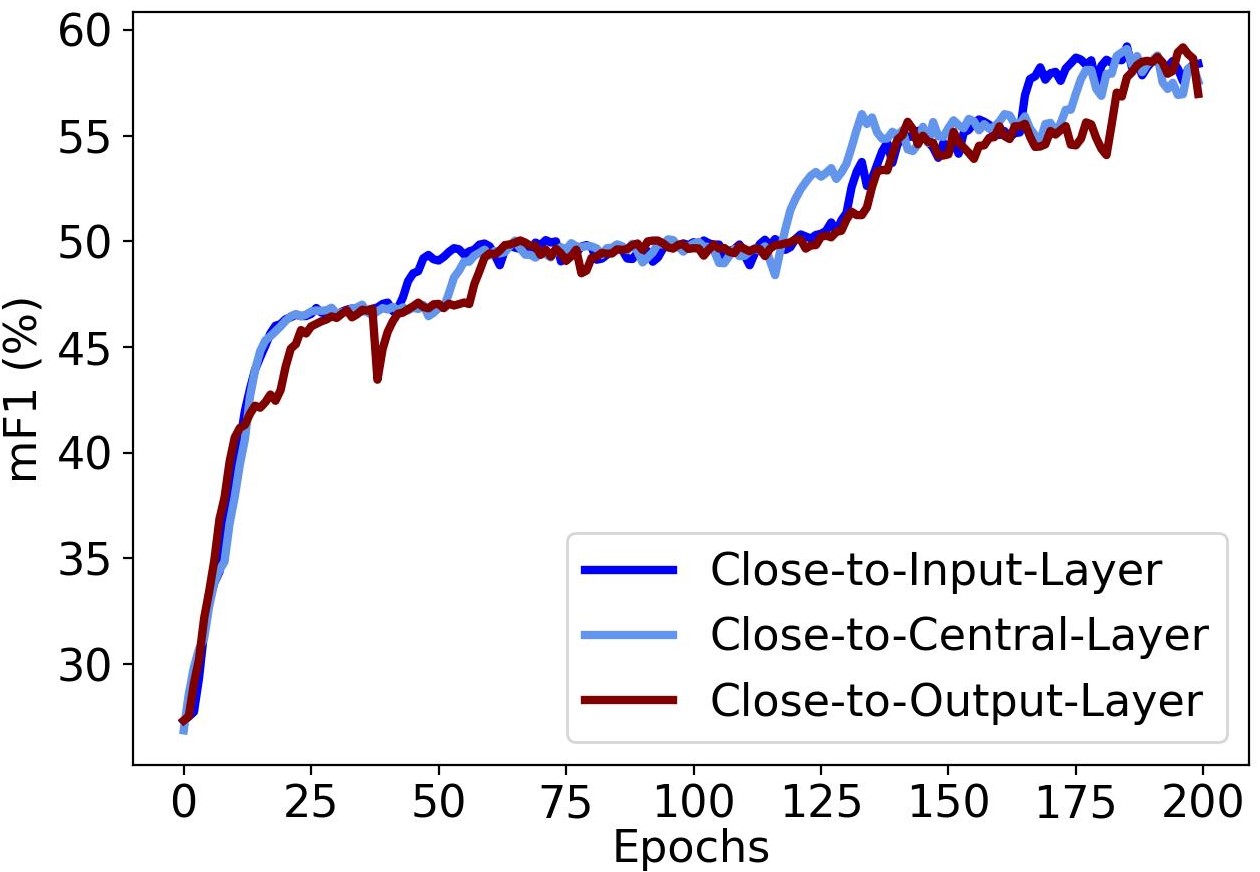 } \\
\hline

\verticaltext[38pt]{\textbf{Three-Point}} &\hspace{-0.21cm}
\includegraphics[width=0.235\linewidth]{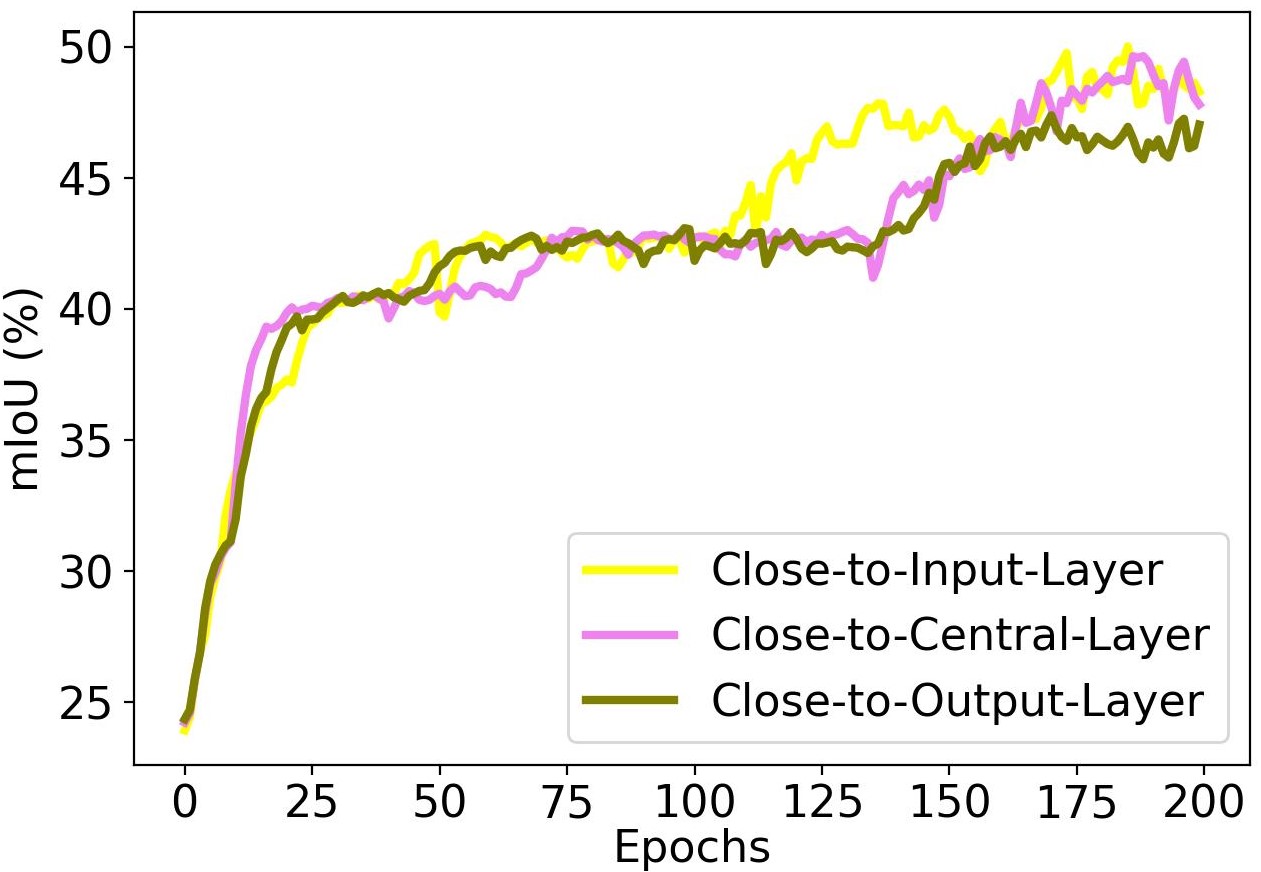} &\hspace{-0.47cm}
\includegraphics[width=0.235\linewidth]{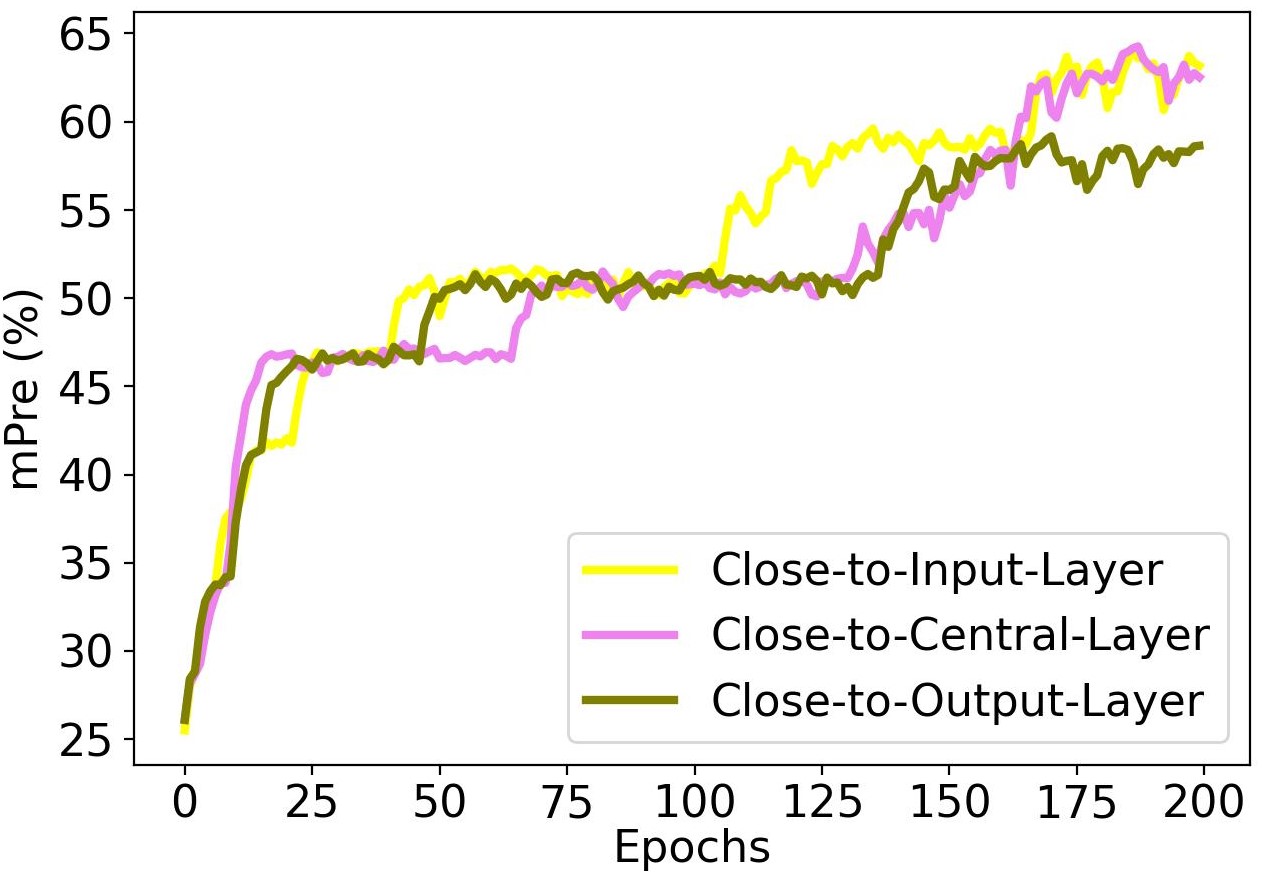} &\hspace{-0.47cm}
\includegraphics[width=0.235\linewidth]{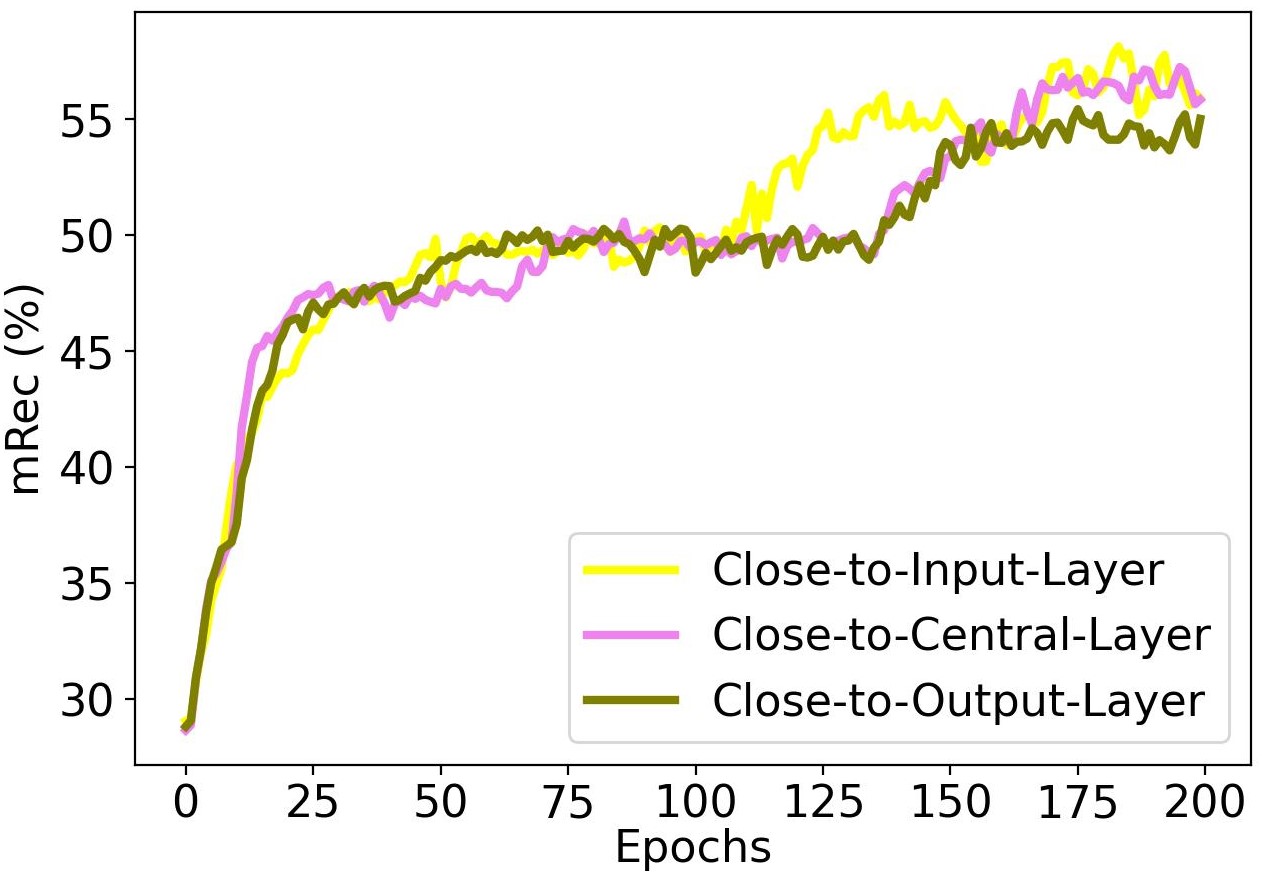} &\hspace{-0.47cm}
\includegraphics[width=0.235\linewidth]{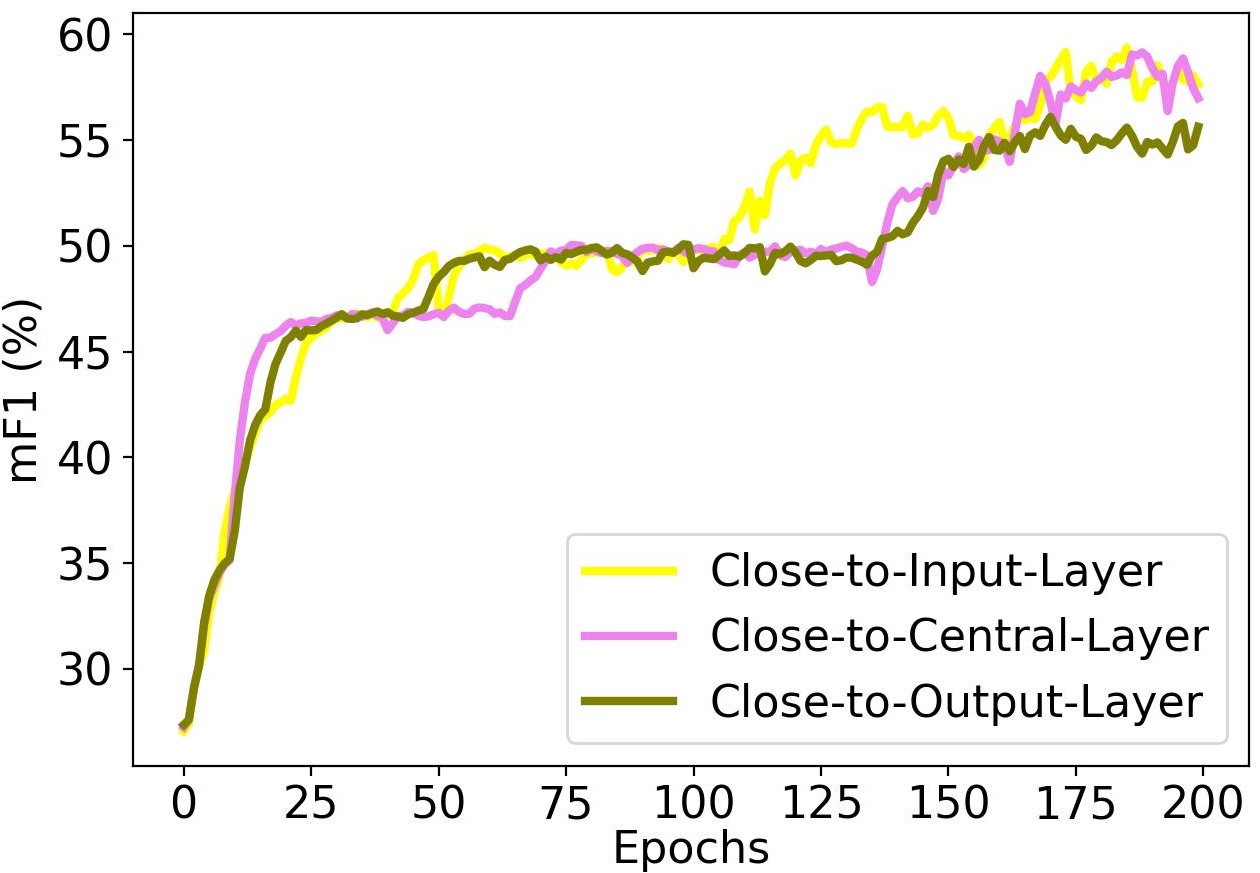 } \\
\hline
\end{tabularx}
\caption{The impact of the position of intermediate points on iMacHSR's training performance.}
\label{tab:abl_position_metric_comp}
\end{table*}

\subsubsection{Qualitative Performance Comparison.}
\Cref{tab:iMacRS_qualitive_comp} illustrates the qualitative performance of the case of enabling iMacHSR against the case of disabling iMacHSR on five RGB images from diverse scenarios. To evaluate the prediction performance of both training methods, we assess how accurately their outputs align with the ground truth and the original images. Our comparison indicates that models trained using iMacHSR consistently deliver superior accuracy, capturing both the broad scene context and intricate details across all images.

\subsection{Ablation Study}
This part reveals three types of ablation study: (I) how the number of intermediate points affects iMacHSR's prediction; (II) how the distance between adjacent intermediate points impacts iMacHSR's prediction; and (III) how positions of intermediate points affects iMacHSR's prediction. 

\subsubsection{The Impact of the Number of Intermediate Points.}
To investigate the role of the number of the intermediate points, we compare five cases with different number of intermediate points ranging from 1 to 5, and they are denoted as ``Intermediate Point (1)'', ``Intermediate Point (2)'', ``Intermediate Point (3)'', ``Intermediate Point (4)'', ``Intermediate Point (5)'', respectively. The comparison results are illustrated in \Cref{Fig.abl_number_Metrics}, from which we can observe that cases with both smaller and larger numbers of intermediate points underperform those with a moderate number. This suggests that in practical training, there is no benefit in setting an excessive number of intermediate points between the input and output layers of the DL model.

\subsubsection{The Impact of the Distance between Adjacent Intermediate Points.}
To figure out how the distance between adjacent intermediate points affects the performance of the proposed iMacHSR, we firstly define the base distance as a fixed number of layers between two adjacent layers. Afterwards, we conduct following three experiments by setting the distance between adjacent intermediate points as (I) one base, (II) two bases, and (III) three bases. The experimental results are illustrated in \Cref{Fig.abl_dist_Metrics}, which indicates that the case with two-base distance achieves the best performance among aforementioned three cases. This inspires us that in training a moderate distance facilitates a better training performance.

\subsubsection{The Impact of the Position of Intermediate Points.}
We conducted three series of experiments to investigate the impact of intermediate point placement in a DL model:
\begin{itemize}
    \item Series I: We place single intermediate point in three distinct positions: close to the input layer, close to the central layer, and close to the output layer.
    \item Series II: We position two intermediate points across the same three locations: close to the input layer, close to the central layer, and close to the output layer.
    \item Series III: We arrange three intermediate points in the aforementioned positions: close to the input layer, close to the central layer, and close to the output layer.
\end{itemize}
The experimental results are revealed in \Cref{tab:abl_position_metric_comp}. From \Cref{tab:abl_position_metric_comp}, we can figure out following common patterns: (I) For each of these three series, the case of ``Close-to-Input-Layer'' consistently outperforms cases of ``Close-to-Central-Layer'' and ``Close-to-Output-Layer''. (II) Across all three series, as the number of intermediate points increases, the performance of the case of ``Close-to-Output-Layer'' progressively deteriorates. (III) Across all three series, increasing the number of intermediate points consistently improves the performance of the case of ``Close-to-Central-Layer''.

\section{Conclusion}
In this study, we address the problem of suboptimal training in DL models due to inadequate supervision for deeper model architectures. We introduce iMacHSR strategy to enhance the DL model optimization. iMacHSR integrates heterogeneous losses for robust intermediate supervision and negative entropy regularization to prevent overconfident predictions. Our experiments demonstrate that iMacHSR effectively improves the performance of DL models across various scenarios, outperforming traditional output-layer supervision method. Future work will refine the supervision weights for optimal training outcomes.

\end{document}